\documentclass[11pt]{article}
\usepackage{psfrag,epsf}
\usepackage{url} 
\newcommand{\blind}{0}

\usepackage[hmargin={0.9in, 0.9in}, vmargin={1in, 1in}]{geometry}

\usepackage{setspace}
\usepackage[utf8]{inputenc}
\usepackage{amsmath, amsfonts, amssymb}
\usepackage{amsthm}
\usepackage{mathtools}
\usepackage{subfigure}
\usepackage{dsfont}
\usepackage{booktabs}
\usepackage{enumerate}
\usepackage{hyperref}
\usepackage[table, dvipsname]{xcolor}
\usepackage{colortbl}
\usepackage{multicol}
\usepackage{multirow}
\usepackage{graphicx}
\usepackage{natbib}
\usepackage[ruled]{algorithm2e}
\usepackage{cleveref}
\usepackage{float}

\usepackage{xr-hyper}
\newtheorem{theorem}{Theorem}[section]
\newtheorem{proposition}[theorem]{Proposition}

\newtheorem{definition}[theorem]{Definition}
\newtheorem{assumption}[theorem]{Assumption}

\newcommand{\eins}{\boldsymbol{1}}
\newcommand{\argmax}{\mathop{\mathrm{argmax}}}

\usepackage{titlesec}
\titlespacing*\section{0pt}{2pt plus 2pt minus 2pt}{0pt plus 2pt minus 2pt}
\titlespacing*\subsection{0pt}{2pt plus 2pt minus 2pt}{0pt plus 2pt minus 2pt}
\titlespacing*\subsubsection{0pt}{0pt plus 2pt minus 2pt}{0pt plus 2pt minus 2pt}

\begin{document}
\setlength{\abovedisplayskip}{2pt}
\setlength{\belowdisplayskip}{2pt}
\setlength{\abovedisplayshortskip}{1pt}
\setlength{\belowdisplayshortskip}{1pt}
\setlength\intextsep{4pt}



\if0\blind
{
\bigskip
\bigskip
\bigskip
\title{\bf Locally Private Nonparametric Contextual Multi-armed Bandits with Transfer Learning}
\author{
Yuheng Ma\thanks{School of Statistics, Renmin University of China, yma@ruc.edu.cn.} \;\;  
Feiyu Jiang\thanks{School of Management, Fudan University, jiangfy@fudan.edu.cn.}  \;\; 
Zifeng Zhao\thanks{Mendoza College of Business, University of Notre Dame, zifeng.zhao@nd.edu.} \;\;
Hanfang Yang\thanks{Center for Applied Statistics, School of Statistics, Renmin University of China, hyang@ruc.edu.cn.}
\;\;
Yi Yu\thanks{Department of Statistics, University of Warwick, yi.yu.2@warwick.ac.uk.} \;\; 
}
\date{}
\maketitle
} \fi

\if1\blind
{
\bigskip
\bigskip
\bigskip
\bigskip
\begin{center}
{\LARGE\bf Locally Private Nonparametric Contextual Multi-armed Bandits}
\end{center}
\medskip
} \fi

\vspace{-1cm}
\begin{abstract}
Motivated by privacy concerns in sequential decision-making on sensitive data, we address the challenging problem of nonparametric contextual multi-armed bandits (MAB) under local differential privacy (LDP). Via a novelly designed LDP-compatible confidence bound, we propose an algorithm that achieves near-optimal regret performance, whose optimality is further supported by a newly derived minimax lower bound. We further consider the case of private transfer learning where auxiliary datasets are available, subject also to (heterogeneous) LDP constraints. Under the widely-used covariate shift framework, we propose a jump-start scheme and a novel reweighted LDP-compatible estimator and confidence bound, which effectively combine and utilize information from heterogeneous auxiliary data. The minimax optimality of the algorithm is further established by a matching lower bound. Comprehensive experiments on both synthetic and real-world datasets validate our theoretical results and underscore the effectiveness of the proposed methods.
\end{abstract}

\noindent%
{\it Keywords:}  local differential privacy, contextual multi-armed bandit, transfer learning, covariate shift

\section{Introduction}
Contextual multi-armed bandit (MAB) \citep[e.g.][]{lu2010contextual, zhou2015survey} is a versatile and general framework for sequential decision-makings and has been widely deployed in various practical domains, such as personalized recommendations \citep[e.g.][]{li2010contextual}, clinical trials \citep[e.g.][]{ameko2020offline}, and portfolio management \citep[e.g.][]{cannelli2023hedging}. 
However, the contextual information in many applications often consists of sensitive user data. For example, clinical trials may include detailed physical and biometric information about patients, while recommendation systems may hold demographics and purchase/view histories information of users. It thus naturally raises privacy concerns given potential data leakage of the sensitive contextual information in MAB.

To address the information security concerns, differential privacy (DP) \citep{dwork2006calibrating} has emerged as the gold standard for protecting user data.  
Depending on the availability of a central server that has access to all information, the notion of DP can be further categorized into central differential privacy (CDP) and local differential privacy (LDP) \citep[e.g.][]{kairouz2014extremal, duchi2018minimax}. 
In the literature, under a parametric assumption on the reward functions, many works have considered private contextual MAB under the CDP setting where a \textit{trusted} central server can store user data \citep[e.g.][]{kusner2015differentially, shariff2018differentially, dubey2020differentially, wang2022dynamic, chakraborty2024fliphat, chen2025near}.  

However, in many practical scenarios, such a trusted central server may not exist and users may prefer to avoid directly sharing any sensitive information with the server. In such cases, LDP serves as an effective privacy-preserving framework. In fact, compared to CDP, LDP is more widely deployed in the industry due to its greater applicability~\citep{erlingsson2014rappor, apple2017differential, tang2017privacy, yang2024local}.
In the literature, contextual MAB has also been studied under the LDP setting \citep[e.g.][]{zheng2020locally, han2021generalized, charisopoulos2023robust, huang2023federated, li2024optimal, zhao2024contextual}, though existing works also primarily focus on parametric reward functions, such as linear and generalized linear models. Indeed, to our knowledge, no prior work has addressed the problem of nonparametric contextual MAB under LDP constraints.



In the era of big data, the decision makers (referred to as server henceforth), such as financial, pharmaceutical and tech companies, often have access to additional data sources (i.e.\ auxiliary data) besides information from the target problem.
This motivates transfer learning (TL) \citep[e.g.][]{cai2021transfer, li2022transfer, cai2022transfer}, a promising area of research in machine learning and statistics, which aims to improve performance in a target domain by leveraging knowledge from related source domains.
Substantial improvement can be achieved via TL when the target and source problems share certain similarities, such as regression function~\citep[e.g.][]{cai2021transfer, pathak2022new} or sparsity structure~\citep[e.g.][]{li2022transfer}.
Importantly, existing works show that TL can effectively leverage auxiliary data and improve regret in both parametric~\citep[e.g.][]{zhang2017transfer} and nonparametric contextual MAB~\citep[e.g.][]{suk2021self,cai2024transfer}, as it can significantly boost the performance of policies in early stages that would otherwise incur high regret. However, with the additional need of preserving privacy, no existing work has investigated private contextual MAB with knowledge transfer.

Identifying these gaps, our work considers contextual MAB under the LDP constraints and aims to address the following three key questions:
\textit{(i) 
What is the fundamental limit of nonparametric contextual MAB under LDP?
(ii)
Can TL with auxiliary data extend this limit?
(iii)
Can effective algorithms be designed to solve contextual MAB with LDP while also incorporating auxiliary data?
}

Our framework allows LDP constraints on both target and auxiliary data. Aligned with the TL literature on contextual MAB~\citep[e.g.][]{suk2021self, cai2024transfer}, we follow the covariate shift framework, where the target and source MAB have the same reward functions but their contextual information may follow different marginal distributions. This setting is suitable when there exists an objectively homogeneous conditional relationship (i.e.\ the reward function) across several parties with population heterogeneity
As a concrete example, the expected outcomes of a clinical trial represent an objective relationship that remains consistent when conditioned on patient features. However, the distribution of patient features may vary across different cooperating medical institutions.

With the aforementioned setup, our contributions are summarized as follows:
\textit{(i)} We formalize the problem of nonparametric contextual MAB under LDP and further extend it to private transfer learning by introducing auxiliary datasets under covariate shift.
\textit{(ii)} We derive minimax lower bounds on the regret, accounting for varying levels of privacy and the extent of covariate shift.
\textit{(iii)} Based on a novelly designed LDP-compatible confidence bound, we propose an efficient policy for LDP contextual MAB, along with a jump-start scheme to further leverage auxiliary data.
\textit{(iv)} We derive a high-probability regret upper bound for the proposed policy, which is near-optimal and matches the minimax lower bound.
\textit{(v)} We conduct extensive numerical experiments on both synthetic and real data to validate our theoretical findings and demonstrate the practical utility of our methodology.

In Section \ref{sec:ldpbandit}, we introduce the problem of nonparametric contextual MAB with LDP and present the proposed methods and theoretical results. We further extend the problem to private TL with auxiliary data in Section \ref{sec:auxiliarydata}. 
Numerical results, including real data applications, and a conclusion with discussions are provided in Sections~\ref{sec:experiments} and~\ref{sec:discussionconclusion}, respectively. 
All technical proofs and detailed descriptions of the numerical experiments are included in the supplement.

\noindent\textbf{Notation.} For any vector $x$, let $x^i$ denote the $i$-th element of $x$. 
For $1 \leq p < \infty$, the $L_p$-norm of $x = (x^1, \ldots, x^d)^{\top}$ is defined by $\|x\|_p := (|x^1|^p + \cdots + |x^d|^p)^{1/p}$.
We use the notation $a_n \lesssim b_n$ and $a_n \gtrsim b_n$ to denote that there exist positive constants $n_1 \in \mathbb{N}$, $c$ and~$c'$ such that $a_n \leq c b_n$ and $a_n \geq c' b_n$, respectively, for all $n \geq  n_1$. 
We denote $a_n\asymp b_n$ if $a_n\lesssim b_n$ and $b_n\lesssim a_n$.
Let $a\vee b = \max (a,b)$ and $a\wedge b = \min (a,b)$. 
For any set $A\subset \mathbb{R}^d$, the diameter of $A$ is defined by $\mathrm{diam}(A):=\sup_{x,x'\in A}\|x-x'\|_2$. 
Let $f_1 \circ f_2$ represent the composition of functions $f_1$ and $f_2$.
Denote the $k$-composition of function $f$ by $f^{\circ k}$.
Let $A\times B$ be the Cartesian product of sets, where $A\in\mathcal{X}_1$ and $B\in \mathcal{X}_2$ for potentially different domains $\mathcal{X}_1$ and~$\mathcal{X}_2$.
For measure $\mathrm{P}$ on $\mathcal{X}_1$ and $\mathrm{Q}$ on $\mathcal{X}_2$, define the product measure $\mathrm{P} \otimes \mathrm{Q}$ on $\mathcal{X}_1\times \mathcal{X}_2$ as $\mathrm{P} \otimes \mathrm{Q} (A\times B)= \mathrm{P}(A) \mathrm{Q}(B)$. 
For a positive integer $k$, denote the $k$-fold product measure on $\mathcal{X}_1^k$ as $\mathrm{P}^k$.
Let the standard Laplace random variable have probability density function $e^{-|x|} / 2$ for $x\in \mathbb{R}$.  
Let $\operatorname{Unif}(\mathcal{X})$ be the uniform distribution over any domain $\mathcal{X}$.  
A ball whose center and radius are $x$ and $r\in(0,+\infty)$, respectively, is denoted as $B(x,r)$. Denote $[K]=\{1,2,\ldots,K\}$ and $[0] = \varnothing$.

\section{Locally Private Nonparametric Contextual Bandits}\label{sec:ldpbandit}
\subsection{Preliminaries}\label{sec:preliminary}

\noindent\textbf{Privacy.} We first rigorously define the notion of LDP.
\begin{definition}[Local Differential Privacy]\label{def:ldp}
Given data $\{Z_i \}_{i=1}^{n} \subset \mathcal{Z}$, a mechanism $\tilde{\mathrm{P}}: \mathcal{Z}^n \to \tilde{\mathcal{Z}}^n$ is sequentially-interactive $\varepsilon$-locally differentially private ($\varepsilon$-LDP) for some $\varepsilon>0$ if,
\begin{align*}
\frac{\tilde{\mathrm{P}}\left(\tilde{Z}_i \in S \mid Z_i = z,  \tilde{Z}_{1}, \ldots, \tilde{Z}_{i - 1} \right)}{\tilde{\mathrm{P}}\left(\tilde{Z}_i \in S \mid Z_i = z',  \tilde{Z}_{1}, \ldots, \tilde{Z}_{i - 1} \right)}  \leq e^{\varepsilon},
\end{align*}
for all $1 \leq i \leq  n, S \in \sigma(\mathcal{\tilde{Z}}), z, z' \in \mathcal{Z}$, and $\tilde{Z}_{1}, \ldots, \tilde{Z}_{i - 1} \in \mathcal{\tilde{Z}}$, where $\tilde{\mathcal{Z}}$ is the space of the outcome.
\end{definition}

This LDP formulation is widely adopted \citep[e.g.][]{duchi2018minimax}, with the statistical procedure operating based only on the private data $\tilde{Z}_1,\ldots,\tilde{Z}_n$. 
The term \textit{sequentially interactive} refers to the privacy mechanisms having access to the privatized historical data, which is particularly suitable for describing the sequential nature of bandit problems. 

\noindent\textbf{Contextual multi-armed bandits.} Let domain $\mathcal{X} = [0,1]^d$, number of arms $K \in \mathbb{Z}_+$ and $\mathrm{P}$ be a probability measure supported on $\mathcal{X}\times[0,1]^{K}$, generating $(X^{\mathrm{P}}, 
Y^{\mathrm{P},(1)}, 
\ldots, Y^{\mathrm{P},(K)})$.
Denote the time horizon by $[n_{\mathrm{P}}]$.
At time $t \in [n_{\mathrm{P}}]$ (i.e.\ for the $t$-th user), based on the covariate $X_{t}^{\mathrm{P}}\in\mathcal{X}$ drawn from the marginal distribution $\mathrm{P}_{X}$, an arm $k\in[K]$ is selected and one receives a random reward $Y_{t}^{\mathrm{P},(k)} \in [0, 1]$ associated with the chosen $k$, whose value is drawn according to the conditional distribution $\mathrm{P}_{Y^{\mathrm{P},(k)}| X_{t}^{\mathrm{P}}}$.
Given $X_{t}^{\mathrm{P}}$, let the conditional expectation of $Y_{t}^{\mathrm{P},(k)}$ be 
\begin{align*}
\mathbb{E}\big[Y_{t}^{\mathrm{P},(k)}\,|\,X_{t}^{\mathrm{P}}\big]=f_{k}(X_{t}^{\mathrm{P}}),
\end{align*}
where $f_{k}:\mathcal{X}\rightarrow[0,1]$ is an unknown reward function associated with arm $k$. Under LDP, the raw information $Z_t^{\mathrm{P}} = (X_t^{\mathrm{P}}, k, Y_t^{\mathrm{P}, (k)})$ of user $t$ needs to be privatized into $\tilde{Z}_t^{\mathrm{P}}$.
For each $t$, define the natural filtration generated by the raw context, arm and reward as $\mathcal{{F}}_{t} := \sigma({Z}_1^{\mathrm{P}}, \ldots, {Z}_t^{\mathrm{P}})$, and define the natural filtration generated by the privatized data as $\mathcal{\tilde{F}}_{t} := \sigma(\tilde{Z}_1^{\mathrm{P}}, \ldots, \tilde{Z}_t^{\mathrm{P}})$. 
Note that~$\tilde{Z}_t^{\mathrm{P}}$ is a function of both ${Z}_t^{\mathrm{P}}$ and $\mathcal{\tilde{F}}_{t-1}$. 

\begin{figure}
\centering
\includegraphics[width = 0.75\textwidth]{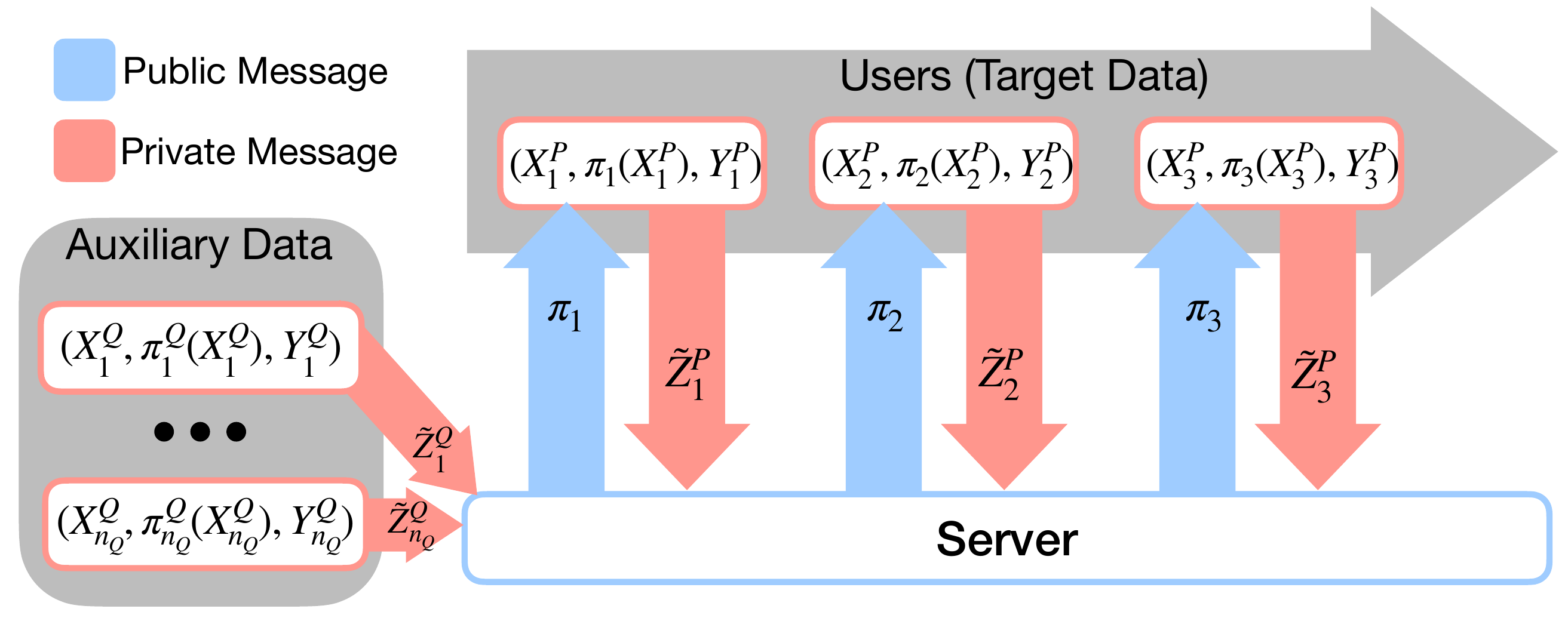}
\caption{Illustration of the learning process. To achieve LDP, the server only receives privatized information $\tilde{Z}^{\mathrm{P}}_t$, while the context $X_t^{\mathrm{P}}$, the pulled arm $\pi_t(X_t^{\mathrm{P}})$, and the reward $Y_t^{\mathrm{P}}$ remains at the user end. The same applies to the auxiliary data.}
\label{fig:privacyillustration}
\end{figure}

A policy $\pi$ is a collection of functions $\{\pi_{t}\}_{t \geq 1}$ where $\pi_{t}: {X}_t^{\mathrm{P}} \times \mathcal{\tilde{F}}_{t - 1}\mapsto [K]$ 
prescribes the policy on choosing which arm to pull at time $t$.
Without confusion, we omit $\mathcal{\tilde{F}}_{t}$ and write the pulled arm by $\pi_t(X_t^{\mathrm{P}})$.
For $\varepsilon > 0$, let $\Pi(\varepsilon)$ be the class of policies that receive information from $\mathcal{D}^{\mathrm{P}} = \{ Z_i^{\mathrm{P}} \}_{i= 1}^{n_{ \mathrm{P} } }$ through an $\varepsilon$-LDP mechanism.
The overall interaction process is illustrated in Figure \ref{fig:privacyillustration}, where we remark that, by design, the sensitive user information $Z_t^{\mathrm{P}}$ always stays on the user side and can only be passed to the server after privatization, and thus achieving LDP.

Let $\pi^{*}$ denote the oracle optimal policy with access to full knowledge of the reward functions $\{f_{k}\}_{k=1}^{K}$, namely $\pi^{*}(x)\in\argmax_{k\in[K]}f_{k}(x)$. 
Our main objective is to design a LDP-preserving policy $\pi \in \Pi(\varepsilon)$ minimizing the regret defined as
\begin{align}
R_{n_{\mathrm{P}}}(\pi)  = \sum_{t=1}^{n_{\mathrm{P}}} \mathbb{E}_{X \sim \mathrm{P}_X}\left[ f_{\pi^{*}(X)}(X)-f_{\pi_{t}(X)}(X) \mid \mathcal{\tilde{F}}_{t - 1} \right].   \label{equ:def-regret}
\end{align}
We remark that in \eqref{equ:def-regret}, each summand is an instant (expected) regret of policy $\pi_t$, where the expectation is taken with respect to the context $X$ that is independent of $\mathcal{\tilde{F}}_{t - 1}$.

\subsection{Minimax Optimal Regret Bound}\label{sec:minimaxrate}

In this section, we investigate the minimax optimal rate of the regret in the problems of contextual MAB subject to LDP. 
The rate is materialized through a lower bound in \Cref{thm:lower-bound} and an upper bound in \Cref{thm:upper-bound}.  The specific class of distributions considered is denoted by $\Lambda(K, \beta)$, i.e.
\begin{align} 
\Lambda& (K, \beta) = \big\{\mathrm{P}\mid \mathrm{P}  \text{ is a distribution supported on $\mathcal{X} \times [0, 1]^K$} \nonumber \\
&\text{satisfying Assumptions \ref{def:smooth} and \ref{def:bounded-density}, and \Cref{def:margin} with parameter $\beta > 0$} \big\}. \label{equ:defofclasslamda}
\end{align}

\begin{assumption}[Smoothness]
\label{def:smooth}
The reward functions $\{f_{k}\}_{k=1}^{K}$ are Lipschitz continuous, i.e.~there exists an absolute constant $C_{L}>0$ such that $$\big|f_{k}(x)-f_{k}(x')\big|\leq C_{L} \| x-x'\|_{2},\text{ for all } x,x'\in\mathcal{X} \text{ and } k \in [K].$$
\end{assumption}

\begin{assumption}[Bounded density]\label{def:bounded-density} 
The marginal density $\mathrm{P}_X$ is bounded, i.e.~there exist absolute constants $\overline{c}>\underline{c}>0$ such that $\underline{c} r^d \leq \mathrm{P}_{X}(B(x,r)) \leq \overline{c} r^d$ for any $x\in \mathcal{X}$ and $r\in(0,1]$.
\end{assumption}

Let $f_{(1)}$ and $f_{(2)}$ denote the pointwise maximum and second maximum functions respectively, namely $f_{(1)}(x)  \coloneqq\max_{k\in[K]}f_{k}(x)$ and
\begin{align*}
f_{(2)}(x)\coloneqq\begin{cases}
\max\limits_{k\in[K]}\big\{ f_{k}(x):f_{k}(x)<f_{(1)}(x)\big\}, & \min\limits_{k\in[K]}f_{k}(x)\neq\max\limits_{k\in[K]}f_{k}(x),\\
f_{(1)}(x), & \text{otherwise.}
\end{cases}
\end{align*}

\begin{assumption}[Margin]\label{def:margin}
The reward functions $\{f_{k}\}_{k=1}^{K}$ satisfy the margin condition, i.e.~there exist absolute constants $\beta, C_{\beta}>0$ such that 
\begin{align*}
\mathbb{P}_{X \sim \mathrm{P}_X}\big(0<f_{(1)}(X)-f_{(2)}(X)\leq\Delta\big)\leq C_{\beta}\Delta^{\beta},\quad \forall \;0<\Delta\leq1.
\end{align*}
\end{assumption}

Assumptions~\ref{def:smooth} and \ref{def:bounded-density} are standard in the nonparametric statistics literature \citep[e.g.][]{audibert2007fast, samworth2012optimal, chaudhuri2014rates}. 
\Cref{def:margin} upper bounds the probability of the event where the best arm is hard to distinguish.  The larger $\beta$ is, the larger the separation 
and hence the easier the problem. 
This characterization of the difficulty of the problem is widely used in the bandit literature \cite[e.g.][]{rigollet2010nonparametric, perchet2013multi, suk2021self, cai2024transfer}.
As noted by \cite{perchet2013multi}, when $\beta > d$—that is, when the separation is excessively large—one of the arms becomes uniformly dominant across $\mathcal{X}$. 
The problem then reduces to a static MAB, which is not our focus.
Consequently, we only consider $\beta \leq d$.

\begin{theorem}[Lower bound]\label{thm:lower-bound}
Consider the class of distributions $\Lambda(K, \beta)$ in \eqref{equ:defofclasslamda} and the class of LDP policies $\Pi(\varepsilon)$. 
It holds that
\begin{align}\label{equ:lowerbound-maintext1}
\inf_{\pi \in \Pi(\varepsilon)}\sup_{\substack{\Lambda(K,\beta )}}\mathbb{E}[R_{n_\mathrm{P}}(\pi)]
\geq  c n_{\mathrm{P}} \Big\{  n_{\mathrm{P}} (e^\varepsilon - 1)^2  \wedge {n_{\mathrm{P}}}^{\frac{2 + 2d}{2 + d}}\Big\}^{-\frac{1+\beta}{2 + 2d }}, 
\end{align}
where $c>0$ is an absolute constant depending only on $d$, $C_L$ and $\beta$.
In particular, when $0< \varepsilon \leq 1$, it holds with an absolute constant $c' > 0$ that
\begin{align}\label{equ:lowerbound-maintext2}
\inf_{\pi \in \Pi(\varepsilon)} \sup_{\substack{\Lambda(K,\beta )}}\mathbb{E}[R_{n_\mathrm{P}}(\pi)]
\geq  c' n_{\mathrm{P}} \left(  n_{\mathrm{P}} \varepsilon^2  \right)^{-\frac{1+\beta}{2 + 2d }}.
\end{align}
\end{theorem}

The proof of Theorem \ref{thm:lower-bound} can be found in Section \ref{sec:Proof-of-Lower-Bound} of the supplement. To accompany the lower bound, in the following, we further present a high-probability upper bound on the regret, which can be achieved by a novel nonparametric LDP bandit algorithm proposed in Section \ref{sec:methodologyP} (\Cref{alg:ldpmabofP}) later.
The proof of \Cref{thm:upper-bound} is provided in Section~\ref{sec:proof-upper-bound}.

\begin{theorem}[Upper bound]\label{thm:upper-bound}
Consider the class of distributions $\Lambda(K, \beta)$ in \eqref{equ:defofclasslamda} and the class of LDP policies $\Pi(\varepsilon)$. 
Suppose $\mathrm{P} \in \Lambda(K,\beta )$. 
Then, we have that the policy $\pi$ given by \Cref{alg:ldpmabofP} satisfies $\pi \in \Pi(\varepsilon)$ and with probability at least $1 -  n_{\mathrm{P}}^{-2}$, 
\begin{align}
R_{n_{\mathrm{P}}}(\pi) \leq C  n_{\mathrm{P}} \Bigg\{ \left(\frac{n_{\mathrm{P}} \varepsilon^2}{K^2\log (n_{\mathrm{P}})} \right) \wedge \left(\frac{n_{\mathrm{P}}}{K \log (n_{\mathrm{P}}) }\right)^{\frac{2 + 2d}{2 + d}}\Bigg\}^{-\frac{1+\beta}{2 + 2d }},
\label{equ:upper-bound}
\end{align}
where $C>0$ is an absolute constant depending only on $d$, $C_L$ and $\beta$.
If in addition that $0< \varepsilon \leq 1$, then it holds with an absolute constant $C'>0$ that
\begin{align}
R_{n_{\mathrm{P}}}(\pi) \leq C'  n_{\mathrm{P}} \left(   \frac{n_{\mathrm{P}} \varepsilon^2}{K^2\log (n_{\mathrm{P}})} \right)^{-\frac{1+\beta}{2 + 2d }}.
\label{equ:upper-bound-2}
\end{align}
\end{theorem}

We first compare Theorems~\ref{thm:lower-bound} and \ref{thm:upper-bound} for the case widely encountered in practice, where
the number of arms $K = O(1)$.  Up to logarithmic factors, in the challenging, high-privacy regime $\varepsilon \in (0, 1]$, Theorems~\ref{thm:lower-bound} and \ref{thm:upper-bound} together lead to the minimax rate for the regret    
\begin{align}\label{equ:simplifiedrate}
n_{\mathrm{P}} \left\{\left(  n_{\mathrm{P}} \varepsilon^2  \right)^{-\frac{1+\beta}{2 + 2d }} \vee n_{\mathrm{P}}^{-\frac{1+\beta}{2 + d}}\right\} = n_{\mathrm{P}}\left(  n_{\mathrm{P}} \varepsilon^2  \right)^{-\frac{1+\beta}{2 + 2d }}.
\end{align} 
The regret in \eqref{equ:simplifiedrate} is a decreasing function of both $\varepsilon$ and $\beta$, which is intuitive as larger $\varepsilon$ and $\beta$ correspond to an easier problem. 
Observing the left-hand side of \eqref{equ:simplifiedrate}, the two terms correspond to private and non-private rates, where the private rate always dominates with $\varepsilon \in (0, 1]$. 

We now provide a detailed discussion on the private and non-private rates in \eqref{equ:simplifiedrate}. 
The non-private term is $n_\mathrm{P}^{1-\frac{1+\beta}{2+d}}$, consistent with the standard rate for nonparametric contextual MAB under Lipschitz continuity \citep[e.g.][]{perchet2013multi, suk2021self, cai2024transfer}. 
As for the private term in~\eqref{equ:simplifiedrate}, the average regret over $n_{\mathrm{P}}$ target data is $\left(  n_{\mathrm{P}} \varepsilon^2  \right)^{-\frac{1+\beta}{2 + 2d }}$, aligning with known convergence rates for generalization error of nonparametric classification under LDP constraints \citep{berrett2019classification}.
Compared to the non-private average regret, which is $n_\mathrm{P}^{-\frac{1+\beta}{2+d}}$, the LDP rate suffers an extra factor of $d$ in the exponent, thus exhibiting a more severe curse of dimensionality—an effect commonly observed in previous LDP studies \citep{berrett2021strongly, sart2021density, gyorfi2022rate}.

We conclude this subsection with discussions regarding the gap between the upper and lower bounds in terms of the logarithmic factors, the number of arms $K$ and the privacy budget $\varepsilon$.
The additional logarithmic term arises due to the high probability argument we use. 
As for $K$, the upper bound \eqref{equ:upper-bound} depends on $K$, while the lower bound \eqref{equ:lowerbound-maintext1} does not.
Such disagreement between the upper and lower bounds
in terms of $K$ is also observed in the literature for non-private nonparametric MAB~\citep[e.g.][]{perchet2013multi, suk2021self}. Note that in practice, the number of arms $K$ is typically fixed, which makes this gap less relevant. A more refined analysis on closing the gap regarding $K$ remains a challenging open problem.
For moderate $\varepsilon$, there is a gap between $e^{\varepsilon} - 1$ dependence in the lower bound \eqref{equ:lowerbound-maintext1} and~$\varepsilon$ in the upper bound \eqref{equ:upper-bound}.  We conjecture that the lower bound is sharp and a different policy is needed to match it.  Such phenomenon is commonly observed in the LDP literature \citep{gyorfi2022rate, xu2023binary, ma2024optimal}, with rates in the moderate $\varepsilon$ regime only studied in the simple hypothesis testing setting \citep[e.g.][]{pensia2023simple}. 

\subsection{Upper Bound Methodology}\label{sec:methodologyP}
\subsubsection{Overview}

To start, we first provide an overview of our proposed method (see the detailed procedure in \Cref{alg:ldpmabofP} later) in this subsection. Due to the nonparametric nature of the problem, we dynamically partition the covariate space $\mathcal{X}$ into a set of hypercubes (i.e.\ bins) and employ a locally constant estimator, subject to LDP, of the reward functions.
The partition strategy converts the contextual problem into a collection of static MAB decision problems, which are then dealt with via a confidence bound based arm elimination procedure. In particular, given that all arms are pulled sufficiently, we can identify and eliminate sub-optimal arms based on local estimates and the corresponding confidence bounds. Furthermore, to ensure the approximation error due to binning is negligible, the partition is dynamically updated via a refinement procedure. 

The main structure of our algorithm is inspired by the adaptive binning and successive elimination~(ABSE) procedure proposed in \cite{perchet2013multi} for non-private nonparametric contextual MAB. However, to accommodate the LDP constraints, substantial modifications are needed on the design of the mechanism for user-server information separation, and on the construction of the nonparametric reward function estimator and its confidence bound. We refer to \Cref{subsec:comparison_other_alg} for a more detailed comparison with \cite{perchet2013multi}.

To proceed, we introduce the policy $\pi_t$ used at time $t \in [n_{\mathrm{P}}]$. 
Specifically, we maintain an active partition $\mathcal{B}_t$, initialized as $\mathcal{B}_1 = \{B_0^1:=[0,1]^d\}$ (i.e.\ the entire covariate domain) and updated dynamically. 
The subscript and superscript of $B_s^j$ denote the depth and index of the bin, respectively, which will be explained in detail later.
For each bin $B_s^{j}$, denote $A_s^{j} \subseteq [K]$ as its active arms set. 
Upon observing a new covariate $X_t^{\mathrm{P}}$, belonging to some $B_s^{j} \in \mathcal{B}_t$, the policy prescribes
\begin{align}\label{equ:partitionbasedpolicy}
\pi_t(X_t^P) = \operatorname{Unif}( A_s^{j}), 
\end{align}
namely selecting an arm uniformly at random from the candidate arm set $A_s^{j}$.

\begin{figure}[!t]
\centering
\subfigure[Estimating reward functions]{
\begin{minipage}{0.28\linewidth}
\centering
\includegraphics[width=\textwidth]{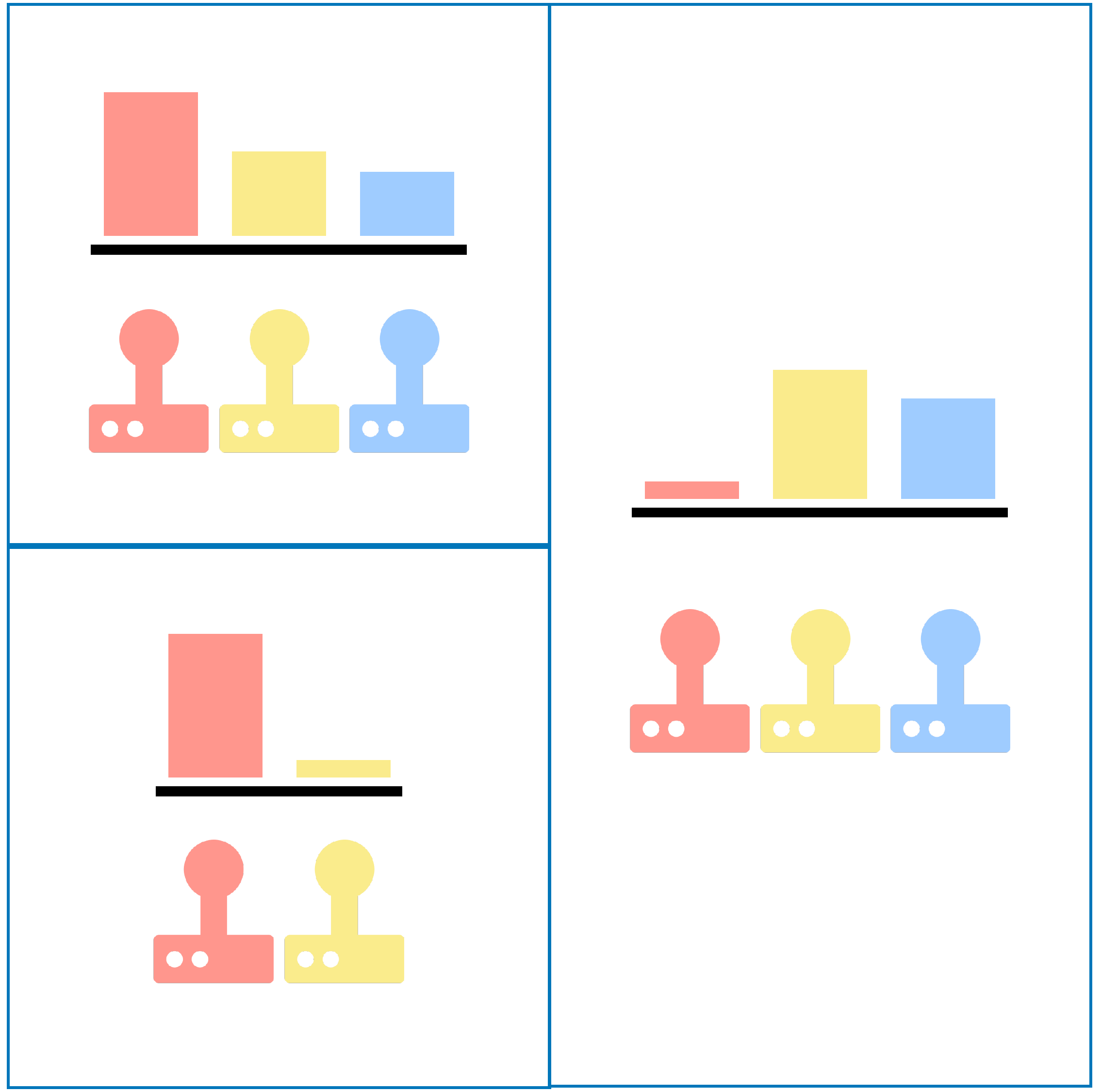}
\end{minipage}
\label{fig:our_algorithm_1}
}
\subfigure[Eliminating arms]{
\begin{minipage}{0.28\linewidth}
\centering
\includegraphics[width=\textwidth]{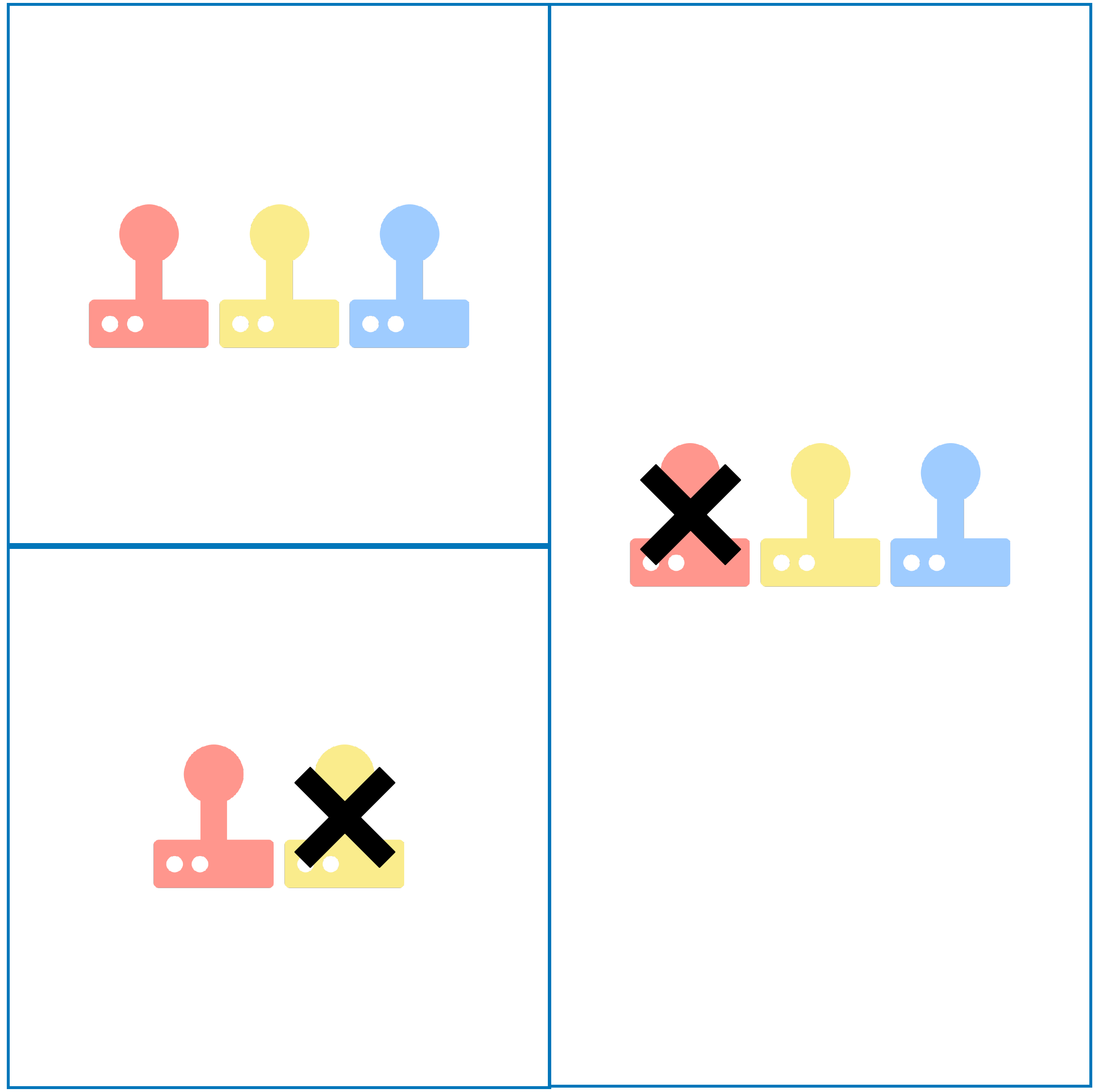}
\end{minipage}
\label{fig:our_algorithm_2}
}
\subfigure[Refining bins]{
\begin{minipage}{0.28\linewidth}
\centering
\includegraphics[width=\textwidth]{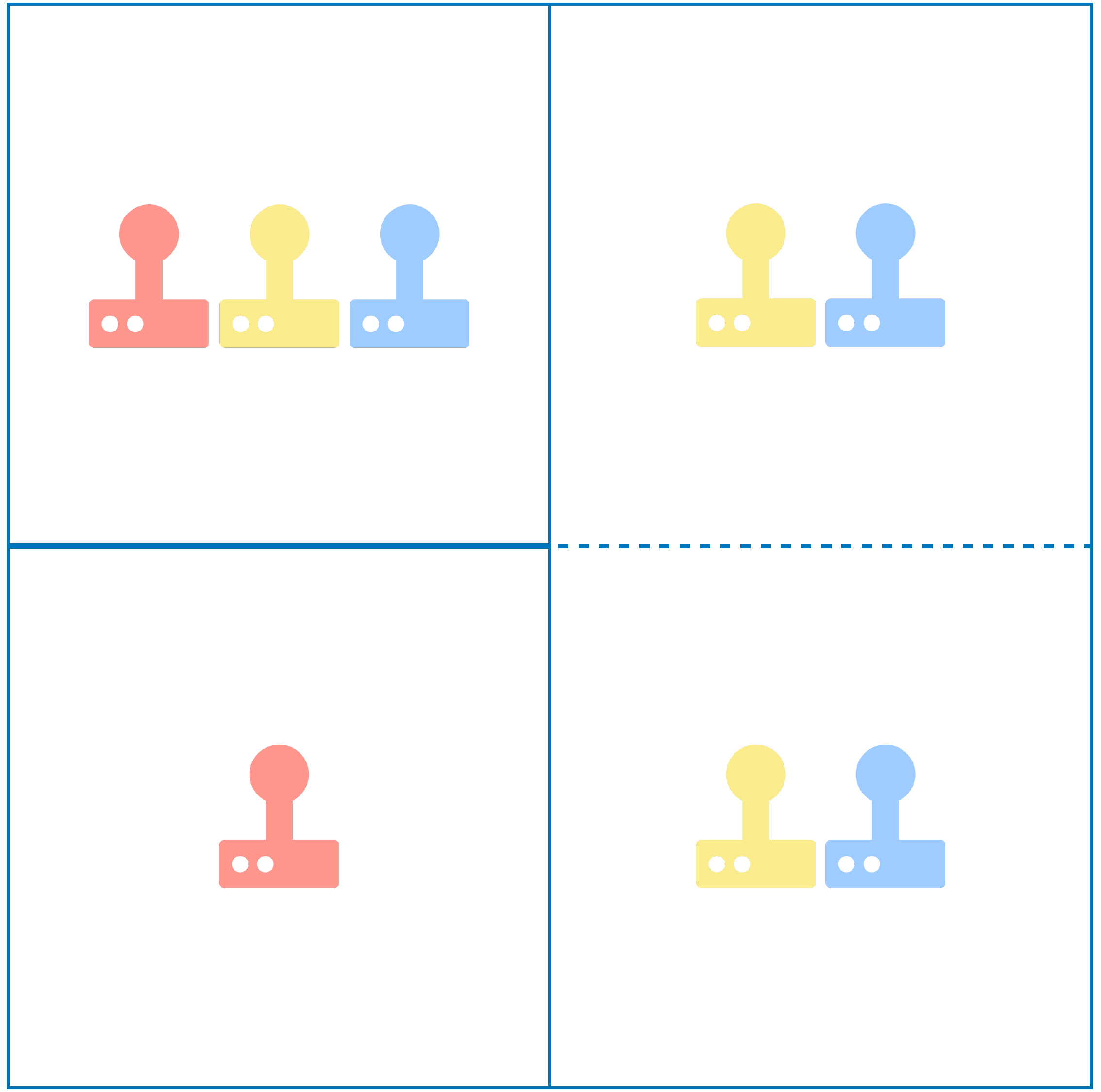}
\end{minipage}
\label{fig:our_algorithm_3}
}
\caption{
Illustration of key steps of the proposed algorithm. 
} 
\label{fig:our_algorithm}
\end{figure}

We now elaborate on how the policy $\pi_t$ is updated across time, which consists of three key components and is further illustrated in \Cref{fig:our_algorithm}. The detailed procedure is given in Algorithm \ref{alg:ldpmabofP}.
\begin{enumerate}
\item Update the private local estimates of the reward functions. 
As shown in Figure \ref{fig:our_algorithm_1}, in each bin, there are $|A_s^{j}|$ active arms, each with its own estimate.
We design a mechanism to optimally estimate the reward functions under LDP. 
This step is formulated in Section \ref{sec:estimatingrewardfunction}. 

\item Decide if any arm needs to be eliminated. Via a novelly constructed confidence bound for LDP nonparametric contextual MAB, we identify and remove suboptimal arms for each bin in the active partition.
This step is illustrated in Figure \ref{fig:our_algorithm_2} and formulated in Section \ref{sec:eliminatingarms}. 

\item Decide whether a given bin should be refined. For any active arm in a given bin, the confidence bound for the local estimate of its reward function becomes narrower as the arm is pulled more times. When the confidence bound is sufficiently narrow, the ability to distinguish sub-optimal arms is restricted by the approximation error of the bin, which can then be improved by refining it to sub-bins. 
This step is illustrated in Figure \ref{fig:our_algorithm_3} and formulated in Section \ref{sec:refiningbins}. 

\end{enumerate}

For clarity of presentation, before discussing the three key components, we first detail the partitioning procedure itself, i.e.~the placement of the dashed lines in \Cref{fig:our_algorithm_3}, in \Cref{sec:partition}.

\subsubsection{Dynamic Partitioning}\label{sec:partition}

\begin{figure}[!t]
\centering
\includegraphics[width = 0.4\textwidth]{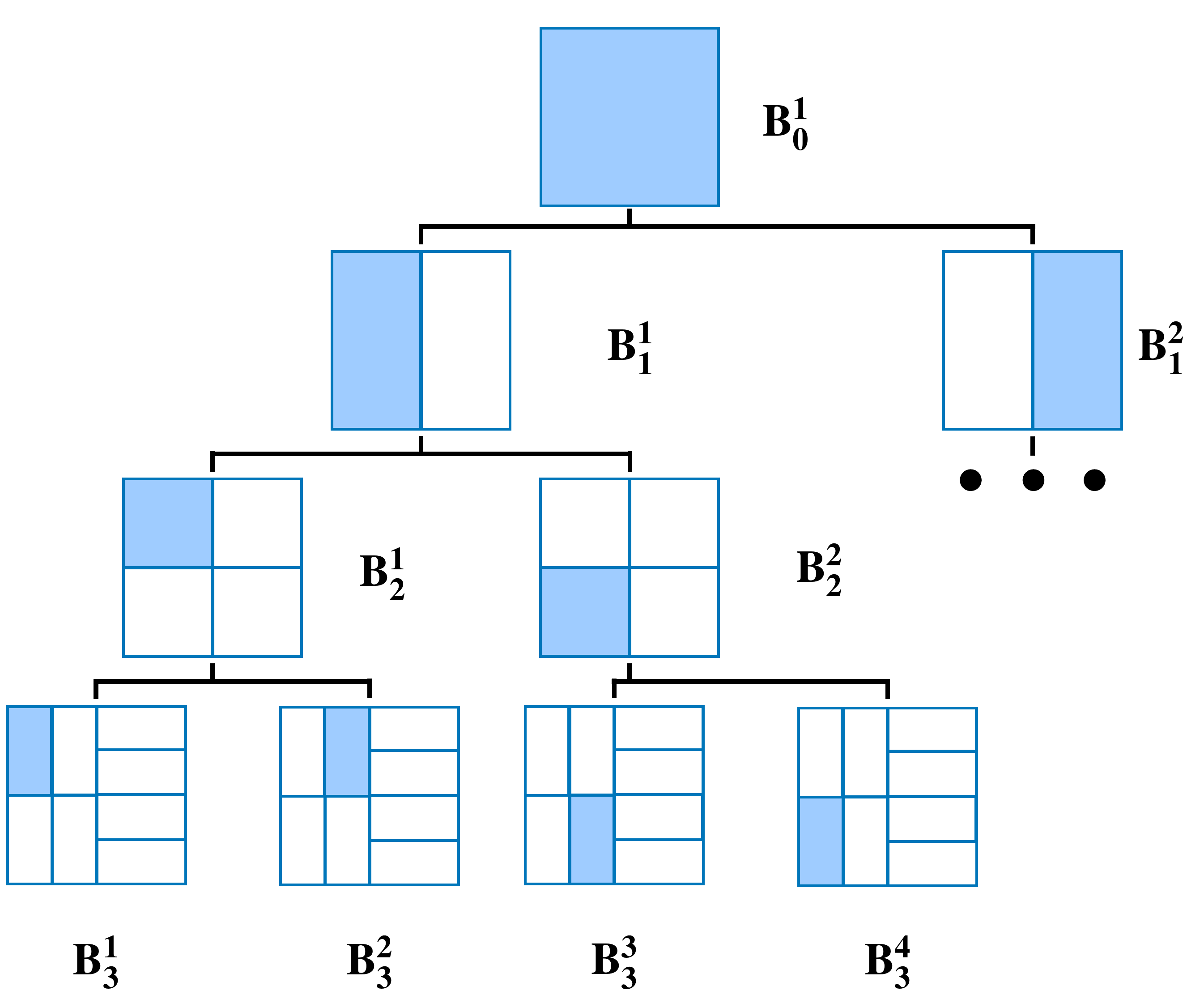}
\caption{A partition created by the max-edge rule for $d = 2$. 
Blue areas give the corresponding bins. 
}
\label{fig:partitionillustration}
\end{figure}

A partition of domain $\mathcal{X}$ is a collection of nonempty, pairwise disjoint subsets whose union is~$\mathcal{X}$.
To create a partition of $\mathcal
X$, let the rectangular bin at the root level be $B_0^0 := [0,1]^d$. 
For each bin $B_s^j$, where $s$ represents its depth and $j \in [2^s]$ is its index, two successive sub-bins are created in the following way. In particular, we uniformly choose a dimension among those embedding longest edges of $B_s^j$, then split $B_s^j$ along this dimension at the midpoint, resulting in sub-bins $B_{s+1}^{2j - 1}$ and $B_{s+1}^{2j}$.   
The partition process is illustrated in Figure \ref{fig:partitionillustration} and formalized in \Cref{alg:partition}. 
This procedure is widely used in the literature and is referred to as dyadic partition or max-edge partition \citep[e.g.][]{blanchard2007optimal, cai2024transfer, ma2025locally}.

\begin{algorithm}[H]
\caption{Max-edge Rule}
\label{alg:partition}
{\bfseries Input: }{Bin $B_{s}^j = \times_{k =1 }^d[a_{sj}^k,b_{sj}^k)$.  } \\
1. Collect $\mathcal{M}_{sj} = \argmax_k |b_{sj}^k - a_{sj}^k| $ and set $k^* =$ Unif $(\mathcal{M}_{sj})$.\\
2. Set $B_{s+1}^{2j-1} = \Big\{x : x\in B_s^j, x^{k^*} < ({a_{sj}^{k^*} + b_{sj}^{k^*}})/{2}\Big\}$ and $B_{s+1}^{2j} = B_s^j / B_{s+1}^{2j-1}$. \\
{\bfseries Output: }{ Sub-bins $B_{s+1}^{2j-1}, B_{s+1}^{2j}$.}
\end{algorithm}

\subsubsection{Estimating Reward Functions}\label{sec:estimatingrewardfunction}

In this section, we study the estimation of reward functions subject to LDP constraints. 
Specifically, we focus on partition-based LDP estimators that assign a constant within each partition bin.
To build intuition, we begin by investigating the non-private counterpart of this partition-based estimation.  It simply averages the rewards of data points whose covariates fall into the same bin.
We then inject the LDP ingredient and present the final estimator.

Let $a_{t, s}^{j} = 1$ if  $B_s^j$ is in $\mathcal{B}_t$ and $0$ otherwise. In other words, $a_{t, s}^{j}$ is the indicator of whether the bin $B_s^j$ is in the active partition $\mathcal{B}_t$ at time $t$. We further define
\begin{align}\label{equ:samplesizebin}
t_s^j = \sum_{i=1}^t a_{i, s}^{j},
\end{align}
which records the total number of times that $B_s^j$ is in the active partition up to time $t.$ Note that both $t_s^j$ and $a_{i, s}^{j}$ are free of privacy concerns since the server is aware of the active partition $\mathcal{B}_t$ at each time step.
Recall the illustration in Figure \ref{fig:privacyillustration}, where $\pi_t$ (and thus its associated active partition $\mathcal{B}_t$) at each step is publicly available. 
In this case, a non-private estimator for $f_k:=f^{\mathrm{P}}_k$ at time $t$ is 
\begin{align}\label{equ:nonprivateestimatorP}
\widehat{f}_{k}^{\mathrm{P}, t} (x)
= 
\sum_{B_s^j \in \mathcal{B}_t}
\eins(x \in B^j_s  ) 
\frac{\sum_{i=1}^{t}Y_i^{\mathrm{P}, (\pi_{i}(X_i^{\mathrm{P}}))} 
\eins(X_i^{\mathrm{P}} \in B^j_s) 
\eins(\pi_{i}(X_i^{\mathrm{P}}) = k) a_{i,s}^{j}}
{\sum_{i=1}^{t}\eins(X_i^{\mathrm{P}} \in B^j_s)\eins(\pi_{i}(X_i^{\mathrm{P}}) = k)a_{i,s}^{j}},
\end{align}
which is simply the sample average of all rewards~(i.e.\ $Y$) that come from arm $k$ with their covariates falling into the same bin in $\mathcal{B}_t$. 
Henceforth, we define $0/0=0$.


For privacy protection, we estimate the reward function under LDP via the Laplace mechanism \citep{dwork2006calibrating}.
Specifically, there are three components in \eqref{equ:nonprivateestimatorP}, namely $Y_i^{\mathrm{P}, (\pi_{i}(X_i^{\mathrm{P}}))}$, $\eins(X_i^{\mathrm{P}} \in B^j_s)$ and $\eins(\pi_{i}(X_i^{\mathrm{P}}) = k)$, that require privatization.
We denote the non-private information at time $i$ by 
\begin{equation}
\label{equ:nonprivatestatistics}
{V}_{i,k,s}^{\mathrm{P},j} = Y_i^{\mathrm{P}, (\pi_{i}(X_i^{\mathrm{P}}))} 
\eins(X_i^{\mathrm{P}} \in B^j_s) 
\eins(\pi_{i}(X_i^{\mathrm{P}}) = k)  \mbox{ and }
{U}_{i,k,s}^{\mathrm{P},j} = 
\eins(X_i^{\mathrm{P}} \in B^j_s) 
\eins(\pi_{i}(X_i^{\mathrm{P}}) = k),
\end{equation}
for $B_s^j \in \mathcal{B}_i$ and $k\in [K]$. 
Specifically, they are privatized as
\begin{align}
\label{equ:privacymechanism}
\tilde{V}_{i,k,s}^{\mathrm{P},j} =  {V}_{i,k,s}^{\mathrm{P},j} + \frac{4  }{\varepsilon} \xi_{i,k,s}^{\mathrm{P},j} \quad \mbox{and} \quad
\tilde{U}_{i,k,s}^{\mathrm{P},j} = 
{U}_{i,k,s}^{\mathrm{P},j} + \frac{4  }{\varepsilon} \zeta_{i,k,s}^{\mathrm{P},j},
\end{align}
where $\xi$'s and $\zeta$'s are i.i.d.~standard Laplace random variables.
The privacy budget $\varepsilon$ is divided into two parts for privacy preservation on $V$'s and $U$'s, respectively.

We remark that \textit{all} $B_s^j \in \mathcal{B}_i$ receives an update based on $X_i^{\mathrm{P}}$ regardless whether $X_i^{\mathrm{P}}\in B_s^j$ or not. 
Otherwise, bin $B_s^j$ not receiving an update reveals $X_i^{\mathrm{P}}\notin B_s^j$, which is a privacy leakage. 
The final estimator is therefore
\begin{align}\label{equ:privateestimatorP}
\tilde{f}_{k}^{\mathrm{P}, t} (x)
= 
\sum_{B_s^j \in \mathcal{B}_t} \eins(x \in B^j_s  ) 
\frac{\sum_{i=1}^{t}
\tilde{V}_{i,k,s}^{\mathrm{P},j} a_{i,s}^{j}}
{\sum_{i=1}^{t} \tilde{U}_{i,k,s}^{\mathrm{P},j} a_{i,s}^{j}},
\end{align}
which satisfies the $\varepsilon$-LDP constraint, as demonstrated in Proposition \ref{prop:privacyguarantee} of the supplement.

\subsubsection{Eliminating Arms}\label{sec:eliminatingarms}

The proposed policy in \eqref{equ:partitionbasedpolicy} uniformly pulls all active arms in $A_s^j$, which implies that we need to exclude arms with large regret from $A_s^j$.
To achieve this, we dynamically rule out arms that are deemed suboptimal in each bin.
By a suboptimal arm in a given bin, we mean an arm whose reward function is lower than that of another arm for all $x$ in the bin.
Although this is an unobservable population property, it can be inferred using a sufficient condition provided in the following proposition.
This proposition establishes a bound between the private estimator \eqref{equ:privateestimatorP} and its population counterpart $\mathbb{E}_{Y|X, \pi}\left[\widehat{f}^{\mathrm{P},t}_{k}(x) \right]$, defined as 
\begin{align*}
\mathbb{E}_{Y|X, \pi}\left[\widehat{f}^{\mathrm{P},t}_{k}(x) \right] =  
\sum_{B_s^j \in \mathcal{B}_t}
\eins(x \in B^j_s  ) 
\frac{\sum_{i=1}^{t} f_{\pi_{i}(X_i^{\mathrm{P}})} (X_i^{\mathrm{P}}) 
\eins(X_i^{\mathrm{P}} \in B^j_s) 
\eins(\pi_{i}(X_i^{\mathrm{P}}) = k) a_{i,s}^{j}}
{\sum_{i=1}^{t}\eins(X_i^{\mathrm{P}} \in B^j_s)\eins(\pi_{i}(X_i^{\mathrm{P}}) = k)a_{i,s}^{j}}. 
\end{align*}
This result will guide the choice of confidence bound in our arm-elimination procedure.

\begin{proposition}\label{prop:finitedifferencebound}
Let $t_s^j = \sum_{i=1}^t a_{i, s}^{j}$ be defined as in \eqref{equ:samplesizebin}. 
With probability at least $1 - n_{\mathrm{P}}^{-2}$, we have for all $t\in [n_{\mathrm{P}}]$ satisfying $t_s^j \geq \log^2 (n_{\mathrm{P}})$,
it holds that,
\begin{align}\label{equ:rkjofP}
\left|\tilde{f}^{\mathrm{P},t}_{k}(x)  - \mathbb{E}_{Y|X, \pi}\left[\widehat{f}^{\mathrm{P},t}_{k}(x) \right]\right|
\leq  
r_{k,s}^{\mathrm{P},t,j} 
:=\sqrt{\frac{C_{ n_{\mathrm{P}}}\left(\left(\varepsilon^{-2} t_s^j\right)\vee \sum_{i=1}^{t}
\tilde{U}_{i,k,s}^{\mathrm{P},j} a^{j}_{i,s} \right) }{\left(\sum_{i=1}^{t}
\tilde{U}_{i,k,s}^{\mathrm{P},j} a^{j}_{i,s}\right)^2}} 
\end{align}
for all $k \in A_s^j, B_s^j \in \mathcal{B}_t$, and $x\in B^{j}_{s}\in \mathcal{B}_t$, where $C_{n_\mathrm{P}} = c\log(n_{\mathrm{P}})$ with a known absolute constant $c$. 
\end{proposition}

The proof of Proposition \ref{prop:finitedifferencebound} can be found in Section \ref{sec:error-analysis} of the supplement, where we also specify the exact expression of $C_{n_{\mathrm{P}}}$. We remark that \Cref{prop:finitedifferencebound} gives the very first confidence bound result for LDP nonparametric contextual bandits.  
In the numerator of~\eqref{equ:rkjofP}, the two terms correspond to the private and non-private bounds, respectively.
The private term $\varepsilon^{-2} t_s^j$ arises from the sum of Laplace random variables.
As for the non-private term, a more natural form is $\sum_{i=1}^{t} {U}_{i,k,s} ^{\mathrm{P},j} a^{j}_{i,s}$, which, however, is unobservable due to the LDP constraints.
Since our algorithm requires an accessible realization of $ r_{k,s}^{\mathrm{P},t,j}$, we replace this term with its private counterpart $\sum_{i=1}^{t} \tilde{U}_{i,k,s} ^{\mathrm{P},j} a^{j}_{i,s}$.

Note that, unlike standard confidence bounds in the nonparametric bandit literature, which count the number of samples whose covariates lie in a particular bin during a given time period, our construction introduces additional Laplacian randomness to comply with LDP requirements.
In Lemma \ref{lem:orderofsumU}, we theoretically show that this substitution does \textit{not} compromise the effectiveness of the confidence bound, provided that $t_s^j$ is sufficiently large. 
To ensure this condition is met in practice, we require a sufficient exploration criterion when conducting arm elimination (see \eqref{equ:updateruleP} below).

An arm elimination rule can be readily derived from \eqref{equ:rkjofP}. In particular, by the triangle inequality, it holds that $|\tilde{f}_{k}^{\mathrm{P},t}(x) -  f_k(x)| \leq 2 r_{k,s}^{\mathrm{P},t,j}$ for all $x\in B_s^j$ provided that
\begin{align}\label{equ:conditionofapproximationpower}
\sup_{x\in B_s^j}\left|\mathbb{E}_{Y|X,\pi}\left[\widehat{f}^{\mathrm{P},t}_{k}(x) \right] -  f_k(x)\right| \leq r_{k,s}^{\mathrm{P},t,j}. 
\end{align}
Here, we refer to $\sup_{x\in B_s^j}\left|\mathbb{E}_{Y|X,\pi}\left[\widehat{f}^{\mathrm{P},t}_{k}(x) \right] -  f_k(x)\right|$ as the approximation error of bin $B_{s}^j$.

Note that condition \eqref{equ:conditionofapproximationpower} can be ensured by the bin refinement procedure introduced in the next subsection.  
Therefore, we can set $2 r_{k,s}^{\mathrm{P},t,j}$ as the radius of confidence bound of $\tilde{f}_{k}^{\mathrm{P},t}(x)$ and we eliminate an arm when its upper confidence bound is smaller than the lower confidence bound of another arm. 
Formally, we remove arm $k^*$ from~$A_s^j$ if  
there exists $k \in [K]$ such that
\begin{align}
t_s^j \geq \log^2\left(n_{\mathrm{P}} \right)  \;\;\text{ and } \;\; \;\; \tilde{f}_{k}^{\mathrm{P},t}(x) -  2r_{k,s}^{\mathrm{P},t,j}  > \tilde{f}_{k^*}^{\mathrm{P},t}(x)  + 2r_{k^*,s}^{\mathrm{P},t,j},
\label{equ:updateruleP}
\end{align}
where the first condition ensures the bin $B_{s}^j$ is sufficiently explored, as is required in Proposition \ref{prop:finitedifferencebound}.

\subsubsection{Refining Bins}\label{sec:refiningbins}

We now introduce the bin refinement procedure to ensure the claimed condition in \eqref{equ:conditionofapproximationpower} holds, which guarantees that the ability to distinguish sub-optimal arms is not
dominated by the approximation error of $B_s^{j}$. 
In particular, utilizing the Lipschitz property of the reward function, we choose 
\begin{align}\label{equ:choiceoftaus}
{\tau}_{s} =2 \sqrt{d}  2^{-s / d},
\end{align}
as a surrogate for the approximation error.
Note that the approximation error is decreasing with $s$, i.e.~the finer bins have smaller errors. 
In fact, \eqref{equ:choiceoftaus} is the diameter of $B_s^j$ and represents an upper bound on the approximation error up to a constant factor, as shown in Lemma \ref{lem:approximationerror} of the supplement. Thus, if $ r_{k,s}^{\mathrm{P}, t, j} < \tau_{s} $ (i.e.\ the confidence bound is sufficiently narrow), it signals insufficient approximation capability of the current bin $B_s^j$, prompting the refinement of the bin using Algorithm \ref{alg:partition}.



\begin{algorithm}[!t]
\caption{The nonparametric MAB algorithm under LDP}
\label{alg:ldpmabofP}
{\bfseries Input:}{ Budget $\varepsilon$. Total sample $n_{\mathrm{P}}.$ }\\
{\bfseries Initialization: }{ $\pi_1 = \operatorname{Unif}([K])$, $\mathcal{B}_1 = \{B_0^1\} =\{ [0,1]^d\}$, $A_0^1 = [K]$.}\\
\For{$t\in [n_{\mathrm{P}}]$}{
\hspace{-0.4cm}\textsc{\color{blue} User side}:\\
Receive $\pi_t$ from the server. Observe $X_t^{\mathrm{P}}$, pull arm $\pi_t(X_t^{\mathrm{P}})$ and receive $Y_{t}^{\mathrm{P},(\pi_t(X_t^{\mathrm{P}}))}$.\\
\For{$B_s^j\in \mathcal{B}_t$}{
\For{$k\in A_s^j$}{ 
Compute $\tilde{V}_{t,k,s}^{\mathrm{P},j}$ and $\tilde{U}_{t,k,s}^{\mathrm{P},j}$ as in \eqref{equ:privacymechanism} and send to the server.  \hfill{\color{blue} \texttt{\# privatization}}\\
}
}

\hspace{-0.4cm}\textsc{\color{blue} Server side}:\\
\For{$B_s^j\in \mathcal{B}_t$}{
\For{$k\in A_s^j$}{ 
Update estimates $\tilde{f}_{k}^{\mathrm{P}, t}$  as in \eqref{equ:privateestimatorP}. {\color{blue}\hfill\texttt{\# estimating reward functions}}\\
Update confidence bounds as in \eqref{equ:rkjofP}.
}
Remove $k$ from $A_s^j$ if \eqref{equ:updateruleP} holds.  \hfill {\color{blue} \texttt{\# eliminating arms}}\\
\If{ $r_{k,s}^{\mathrm{P}, t, j}< {\tau}_{s}$ \textup{for some} $k \in A_s^j$ }
{Generate $B_{s+1}^{2j-1}, B_{s+1}^{2j}$ from $B_{s}^{j}$ using Algorithm \ref{alg:partition}.
\\
$\mathcal{B}_{t} = \mathcal{B}_{t} \cup \{B_{s+1}^{2j-1}, B_{s+1}^{2j}\} \setminus B_{s}^{j}$.  \hfill{\color{blue} \texttt{\# refining bins}}\\
$A_{s+1}^{2j-1} = A_s^{j}$, $A_{s+1}^{2j} = A_s^{j}$. 
}
}
Set $\mathcal{B}_{t+1}=\mathcal{B}_t$, update $\pi_{t+1}$ by \eqref{equ:partitionbasedpolicy} and send to the next user.
}
\end{algorithm}

\subsubsection{Summary and discussions}\label{subsec:comparison_other_alg}
Putting things together, Algorithm \ref{alg:ldpmabofP} summarizes the detailed procedure of our proposed algorithm.

Our upper bound algorithm offers several advantages. First, it is essentially tuning-free, meaning that no hyperparameter needs to be predetermined. 
Moreover, it is sequentially-interactive: once a user sends the privatized $\tilde{V}_{i,k,s}^{\mathrm{P},j}$  and $\tilde{U}_{i,k,s}^{\mathrm{P},j}$, it can safely exit the system~(e.g.\ websites). 
This property is particularly beneficial in industrial settings since it is challenging to continuously track and communicate with users once they leave the system.
Finally, as shown previously in Section \ref{sec:minimaxrate}, our algorithm achieves the near-optimal regret upper bound. 

As discussed before, our algorithm is inspired by the adaptive binning and successive elimination~(ABSE) algorithm proposed in \cite{perchet2013multi}. Here, we highlight their key differences, which stems from the LDP constraints. First, to preserve privacy, our algorithm separates the user-server operations and only allows privatized information exchange between the two sides. Therefore, it is necessary to design new and efficient private nonparametric reward function estimator and the corresponding confidence bound for our policy, which is more challenging than the non-private setting. Second, without privacy concerns, ABSE has the luxury of being able to access and thus leverage \textit{all} past information for updating its policy, which is not feasible under the LDP constraints. As a concrete example, suppose that  at time $i$, for a given bin $B_s^j$ in the active partition $\mathcal{B}_i$, we query the $i$-th user with a privacy budget $\varepsilon$ to construct $\tilde{U}_{i,k,s}^{\mathrm{P}, j}$. If $B_s^j$ is subsequently refined into sub-bins $B_{s+1}^{2j-1}$ and $B_{s+1}^{2j}$, the raw data of the $i$-th user \textit{cannot} be re-queried to construct $\tilde{U}_{i,k,s+1}^{\mathrm{P}, 2j-1}$ or $\tilde{U}_{i,k,s+1}^{\mathrm{P}, 2j}$ as we have used up the $\varepsilon$ privacy budget. In addition, due to privatization, the $\tilde{U}_{i,k,s}^{\mathrm{P}, j}$ quantity cannot be utilized via post-processing to (approximately) determine which sub-bin the $i$-th user belongs to, rendering it unusable in the subsequent learning process. Indeed, this is why our algorithm designs the indicators $a_{i,s}^j$, which disables past (privatized) information once a bin is refined.

One might suggest querying a user multiple times using privacy composition techniques \citep[e.g.][]{dwork2010boosting}. 
However, this approach would require dividing the already limited privacy budget $\varepsilon$, yielding a loss of efficiency. Moreover, it requires to continuously track and communicate with the users, which is not ideal under industry settings.
Another option would be to create a fixed partition with a pre-determined depth.
Though the fixed partition can collect (privatized) information from all samples, it introduces a highly sensitive hyperparameter, i.e.\ the depth of the partition, the choice of which is not obvious and thus is undesirable in practice.

\section{Auxiliary Data Source: A Jump-start}\label{sec:auxiliarydata}

In this section, we further extend our study to transfer learning~(TL) and discuss how auxiliary data can bring a jump-start effect to the nonparametric contextual MAB under the LDP constraints.

\subsection{Preliminaries}\label{sec:preliminaryauxiliary}

In addition to the target data $\{\tilde{Z}^{\mathrm{P}}_t\}_{t\geq 1}$, which comes in sequentially, we assume that there are $M \in \mathbb{Z}_+$ auxiliary datasets $\mathcal{D}^{\mathrm{Q}_1}, \dots, \mathcal{D}^{\mathrm{Q}_M}$, where $\mathcal{D}^{\mathrm{Q}_m} \coloneqq \{Z^{\mathrm{Q}_m}_i\}_{i=1}^{n_{\mathrm{Q}_m}}$ and $Z^{\mathrm{Q}_m}_i = (X_{i}^{\mathrm{Q}_m}, \pi_{i}^{\mathrm{Q}_m}(X_{i}^{\mathrm{Q}_m})$,  $Y_{i}^{\mathrm{Q}_m,(\pi_{i}^{\mathrm{Q}_m}(X_{i}^{\mathrm{Q}_m}))})$ are generated similarly on $\mathcal{X}\times[0,1]^{K}$ based on policy $\pi^{\mathrm{Q}_m}$. 
For now, we assume that the auxiliary data are historical datasets, meaning that all auxiliary data are ready to be queried before we initiate interaction with the target data.
We discuss in Section \ref{sec:discussionconclusion} the case where the auxiliary data are in the form of streaming data - a scenario conforming to the multi-task learning setting. 
We assume $\pi_{i}^{\mathrm{Q}_m}$'s are fixed behavior policies, i.e. $\pi^{\mathrm{Q}_m}_i\equiv \pi^{\mathrm{Q}_m} $, $m\in [M]$ and $i\in [n_{\mathrm{Q}_m}]$.  
Behavior policy is suitable for describing batched data \citep{lange2012batch, levine2020offline} and is widely used in the literature of MAB with auxiliary data \citep[e.g.][]{zhang2017transfer, cai2024transfer}.

\noindent\textbf{Distribution shift.} We allow differences between the distributions of target and auxiliary data by adopting the covariate shift setting. 
In particular, we allow the marginal distributions of covariates in the $\mathrm{P}$-bandit and $\mathrm{Q}$-bandits to be different (i.e. $\mathrm{P}_{X} \neq \mathrm{Q}_{m, X}$, for all $1 \leq m \leq M$), while the distributions of rewards conditioned on the covariate are assumed to be identical, i.e.~$\mathrm{P}_{Y^{(k)}|X} = \mathrm{Q}_{m, Y^{(k)}|X}$ for all $1\leq k\leq K$ and $1\leq m \leq M$. 
We denote the common reward function of the $k$-th arm as $f_{k}(x)\coloneqq f_{k}^{\mathrm{P}}(x)\equiv f_{k}^{\mathrm{Q}_m}(x)$ for all $k\in[K]$ and $x\in\mathcal{X}$.

\noindent\textbf{Privacy.} We allow the target data policy $\pi$ to receive information from $\mathcal{D}^{\mathrm{Q}_m}$ via a sequentially-interactive $\varepsilon_m$-LDP mechanism.
The privacy budgets $\varepsilon_m$ are allowed to vary across the $M$ auxiliary datasets. 
We denote the class of policies that are $(\varepsilon, \varepsilon_1,\dots, \varepsilon_M)$-LDP with respect to $(\mathcal{D}^{\mathrm{P}}, \mathcal{D}^{\mathrm{Q}_1}, \dots, \mathcal{D}^{\mathrm{Q}_M})$ by $\Pi(\varepsilon, \varepsilon_1, \ldots ,\varepsilon_M)$.

\subsection{Minimax Optimal Regret Bound}

We first characterize the connections and differences between the auxiliary and target distributions through the following assumptions.

\begin{definition}[Transfer exponent]\label{def:transfer}Define the transfer exponent $\gamma_m \geq 0$ of $\mathrm{Q}_{m}$ with respect to $\mathrm{P}$ to be the smallest constant 
such that
\begin{align}
\label{eq:tran-exp}
\mathrm{Q}_{m, X}(B(x, r)) \geq C_{\gamma_m}r^{\gamma_m} \mathrm{P}_{X}(B(x, r)),\quad\forall x\in\mathcal{X}, r \in (0, 1],
\end{align}
for some constant $0 < C_{\gamma_m} \leq 1$. 
Let $\gamma = \left(\gamma_1,\dots, \gamma_m\right)^{\top}$.  
\end{definition}

\begin{definition}[Exploration coefficient]
\label{def:explore-coef}
For $m \in [M]$, let $\pi^{\mathrm{Q}_m}(x) = \mu_m(k \, | \, x)$ be a random function over the arm set $[K]$.
Define the exploration coefficient $ \kappa_m  \in [0, 1]$ as 
\begin{align} 
\label{eq:explore-coef}
\kappa_m  \coloneqq K \cdot \inf_{\substack{k\in [K]}}  \mu_m(k \, | \, x), \;\; \forall x\in \mathcal{X}.
\end{align} 
Let $\kappa = \left(\kappa_1,\dots, \kappa_m\right)^{\top}$.   
\end{definition}

Given Definitions \ref{def:transfer} and \ref{def:explore-coef}, we consider the following class of contextual MABs 
\begin{align}\label{equ:classofmabs}
\Lambda(K,\beta,\gamma,\kappa) : =  \left\{(\mathrm{P},\{\mathrm{Q}_m\}_{m = 1}^{M}) \mid \mathrm{P} \in \Lambda(K,\beta); \eqref{eq:tran-exp}\mbox{ and } \eqref{eq:explore-coef} \mbox{ hold for } \mathrm{Q}_m, \forall m\in [M]\right\}.
\end{align}
We comment on these concepts. The transfer exponent is a widely used term for quantifying covariate shift \citep[e.g.][]{kpotufe2021marginal, cai2024transfer}. It requires that the minimum probability under $\mathrm{Q}$ within a given ball is comparable to that under $\mathrm{P}$. 
Clearly, if $\mathrm{Q}_m = \mathrm{P}$, then $\gamma_m = 0$. 
A larger $\gamma_m$ indicates a greater distribution discrepancy.
Definition \ref{def:explore-coef} pertains to the historical data setting, suggesting that the behavior policies should sufficiently explore all arms.

Based on the assumptions, we first establish a minimax lower bound on the regret in Theorem \ref{thm:lower-bound-auxilary}. 
Accordingly, Theorem \ref{thm:upper-bound-auxilary} provides a nearly matching high-probability upper bound on the regret. 
The proof of Theorems \ref{thm:lower-bound-auxilary} and \ref{thm:upper-bound-auxilary} can be found in Appendices \ref{sec:Proof-of-Lower-Bound} and \ref{sec:proof-upper-bound}, respectively. 

\begin{theorem}[Lower bound]\label{thm:lower-bound-auxilary}
Consider the class of distributions $\Lambda(K,\beta,\gamma,\kappa)$ defined in~\eqref{equ:classofmabs} and the class of LDP policies $\Pi(\varepsilon, \varepsilon_1, \ldots , \varepsilon_M)$. 
It holds that
\begin{align}\nonumber
\inf_{\pi\in \Pi(\varepsilon, \varepsilon_1, \ldots ,\varepsilon_M)} \sup_{\substack{\Lambda(K,\beta,\gamma,\kappa )}} & \mathbb{E}[R_{n_\mathrm{P}}(\pi)]
\geq   c n_{\mathrm{P}} \Bigg[   n_{\mathrm{P}} (e^\varepsilon - 1)^2  \wedge n_{\mathrm{P}}^{\frac{2 + 2d}{2 + d}}
\\ + &\sum_{m = 1}^{M} \left(\frac{\kappa_m^2 n_{\mathrm{Q}_m}}{K^2} (e^{\varepsilon_m} - 1)^2\right)^{\frac{2 + 2 d}{2  + 2d + 2\gamma_m}} \wedge \left(\frac{\kappa_m n_{\mathrm{Q}_m}}{K}\right)^{\frac{2 + 2 d}{2 + d + \gamma_m}}\Bigg]^{-\frac{1+\beta}{2 + 2d }},
\label{equ:lower-bound-auxilary}
\end{align}
where $c>0$ is an absolute constant depending only on $d, C_L, \beta, M, \gamma$. 
\end{theorem}

\Cref{thm:lower-bound-auxilary} indicates that the regret can be improved when auxiliary data is available and it further recovers the lower bound result in \Cref{thm:lower-bound} when setting $M=0$. In the lower bound \eqref{equ:lower-bound-auxilary}, the term associated with the auxiliary data contains a factor of~$K$, while the term associated with the target data does not. This arises from our assumption that the policies that generate the auxiliary data are fixed behavior policies (i.e.\ not adaptively updated over time). 
In addition, note that for the term associated with the auxiliary data in \eqref{equ:lower-bound-auxilary}, the dependencies on the number of arms are $K$ and $K^2$ for its non-private and private components, respectively, suggesting that increasing the number of arms introduces greater challenges under privacy constraints.

\begin{theorem}[Upper bound]\label{thm:upper-bound-auxilary}
Consider the class of distributions $\Lambda(K,\beta,\gamma,\kappa)$ defined in~\eqref{equ:classofmabs} and the class of LDP policies $\Pi(\varepsilon, \varepsilon_1, \ldots ,\varepsilon_M)$. 
Suppose that $(\mathrm{P},\{\mathrm{Q}_m\}_{m = 1}^{M}) \in \Lambda(K,\beta,\gamma,\kappa)$. 
Then, we have that the policy $\pi$ given by Algorithm~\ref{alg:ldpmab} satisfies $\pi \in \Pi(\varepsilon, \varepsilon_1, \ldots ,\varepsilon_M)$ and with probability at least $1 - n^{-2}$, the regret of $\pi$ satisfies that 
\begin{align}\nonumber
R_{n_{\mathrm{P}}}(\pi) \leq  C n_{\mathrm{P}} \Bigg[ &\left(\frac{n_{\mathrm{P}} \varepsilon^2}{K^2 \log (n)} \right) \wedge \left(\frac{n_{\mathrm{P}}}{K \log (n)}\right)^{\frac{2 + 2d}{2 + d}}
\\ +& \sum_{m = 1}^{M} \left(\frac{\kappa_m^2 n_{\mathrm{Q}_m} \varepsilon_m^ 2}{K^2 \log (n)} \right)^{\frac{2 + 2 d}{2 + 2d + 2 \gamma_m}} \wedge \left(\frac{\kappa_m n_{\mathrm{Q}_m}}{K \log (n)}\right)^{\frac{2 + 2 d}{2 + d + \gamma_m}} \Bigg]^{-\frac{1+\beta}{2 + 2d}},
\label{equ:upper-bound-auxilary}
\end{align}
where $C>0$ is an absolute constant depending only on $d, C_L, \beta, M, \gamma$ and $n = n_{\mathrm{P}}\vee (\max_{m=1}^M n_{\mathrm{Q}_m})$ is the maximum sample size.
\end{theorem}

Treating the number of arms $K$ as a constant and considering the challenging, high-privacy regime that $\max\{\varepsilon, \varepsilon_1,\cdots,\varepsilon_M\} \in (0, 1]$, we have that, up to the logarithmic factors, the minimax rate of the regret is of order
\begin{align}\label{equ:simplifiedrateauxiliary}
n_{\mathrm{P}} \left\{
n_{\mathrm{P}} \varepsilon^2  + \sum_{m = 1}^{M} \left(\kappa_m^2 n_{\mathrm{Q}_m} \varepsilon_m^ 2 \right)^{\frac{2 + 2 d}{2 + 2d + 2 \gamma_m}}\right\}^{-\frac{1+\beta}{2 + 2d }}.
\end{align}
Compared to the minimax rate without TL in \eqref{equ:simplifiedrate}, we observe that \eqref{equ:simplifiedrateauxiliary} has an increased effective sample size, showing the benefit of auxiliary data.
The contributions of the auxiliary data, compared to target data, are reduced by a polynomial factor of $\kappa_m$ and an exponential factor of $\gamma_m$, which is indeed intuitive and interpretable. 
When $\kappa_m$ is small, there are arms rarely explored, which could potentially be the best arm, thereby limiting the contributions of the auxiliary datasets. 
When $\gamma_m$ is large, the marginal distribution $\mathrm{Q}_{m,X}$ can deviate significantly from $\mathrm{P}_X$, providing redundant information in regions where it is unnecessary.
This also reduces the effective sample size of $\mathcal{D}^{\mathrm{Q}_m}$.

\subsection{Upper Bound Methodology}

We now demonstrate how to leverage the auxiliary data to enhance the performance of our policy in \Cref{alg:ldpmabofP} by designing an additional jump-start stage, where we apply a similar arm elimination procedure 
starting from $n_{\mathrm{Q}_1}$ samples of $\mathcal{D}^{\mathrm{Q}_1}$, continuing with $n_{\mathrm{Q}_2}$ samples of $\mathcal{D}^{\mathrm{Q}_2}$, and finishing off with the $n_{\mathrm{Q}_M}$ samples of $\mathcal{D}^{\mathrm{Q}_M}$. 
Each sample interacts with the policy only once.
We then proceed with learning on $\mathcal{D}^{\mathrm{P}}$.
Therefore, learning on the target data can utilize the refined partition and the set of the selected active arms learned via the source data. For a concrete illustration of such benefits, see Figure \ref{fig:bandit_process} in the numerical experiments section. 

We proceed by defining some necessary notations. First, we simplify the notation by re-indexing the time indices in each dataset with $t \in [n_{\mathrm{P}} + \sum_{m=1}^Mn_{\mathrm{Q}_m}]$, defined as the total number of users that have interacted with policy $\pi$. We further define  
\begin{align*}
T_m(t) = \begin{cases}
0, & t \leq  \sum_{m'\in [m-1]} n_{\mathrm{Q}_{m'}},\\
t - \sum_{m'\in [m-1]} n_{\mathrm{Q}_{m'}},  &  \sum_{m'\in [m-1]} n_{\mathrm{Q}_{m'}} < t \leq  \sum_{m'\in [m]} n_{\mathrm{Q}_{m'}},\\
n_{\mathrm{Q}_m}, & t > \sum_{m'\in [m]} n_{\mathrm{Q}_{m'}}, 
\end{cases}
\end{align*}
for all $m\in [M]$, which gives the total number of users from the $m$-th auxiliary dataset that have interacted with policy $\pi$ up to time $t.$ Analogously, define the target time index by $T_0(t) = (t - \sum_{m=1}^{M} n_{\mathrm{Q}_m}) \vee 0$.
For notational simplicity, we further denote $\mathrm{P}$ as $\mathrm{Q}_0$ and write
$n_{\mathrm{Q}_0} = n_{\mathrm{P}}$, $\varepsilon_0 = \varepsilon$.
For $m \in [M] \cup \{0\}$, define $a_{i,s}^{m,j} = 1$ if  $Z_i^{\mathrm{Q}_m}$ is used to update $B_s^j$ and $0$ otherwise. 
Let the cumulative sample size be $t_s^{m,j} = \sum_{i=1}^{T_m(t)} a_{i,s}^{m,j}$.
Similar to 
\eqref{equ:nonprivatestatistics} and \eqref{equ:privacymechanism}, we encode the information from the auxiliary data by 
\begin{equation}
\begin{aligned}
\label{equ:nonprivatestatisticsauxilary}
{V}_{T_m(t),k,s}^{\mathrm{Q}_m,j} = & Y_{T_m(t)}^{\mathrm{Q}_m, \big(\pi_{T_m(t)}^{\mathrm{Q}_m}(X_{T_m(t)}^{\mathrm{Q}_m})\big)} 
\eins(X_{T_m(t)}^{\mathrm{Q}_m} \in B^j_s) 
\eins(\pi^{\mathrm{Q}_m}(X_{T_m(t)}^{\mathrm{Q}_m}) = k)  ,  \\
{U}_{T_m(t),k,s}^{\mathrm{Q}_m,j} = & 
\eins(X_{T_m(t)}^{\mathrm{Q}_m} \in B^j_s) 
\eins(\pi^{\mathrm{Q}_m}(X_{T_m(t)}^{\mathrm{Q}_m}) = k) ,
\end{aligned}
\end{equation}
for $t\in[\sum_{m=1}^Mn_{\mathrm{Q}_m}], k\in[K], B_s^j \in \mathcal{B}_t$ and $m\in[M]$. 
They are then privatized as
\begin{align}
\label{equ:privacymechanismauxiliary}
\tilde{V}_{T_m(t),k,s}^{\mathrm{Q}_m,j} =  {V}_{T_m(t),k,s}^{\mathrm{Q}_m,j} + \frac{4 }{\varepsilon_m} \xi_{T_m(t),k,s}^{\mathrm{Q}_m,j} ,  \;\;
\tilde{U}_{T_m(t),k,s}^{\mathrm{Q}_m,j} = 
{U}_{T_m(t),k,s}^{\mathrm{Q}_m,j} + \frac{4  }{\varepsilon_m} \zeta_{T_m(t),k,s}^{\mathrm{Q}_m,j}. 
\end{align}

We present the detailed algorithm for leveraging auxiliary data in Algorithm~\ref{alg:ldpmab}. 
The algorithm essentially repeats the sequential procedures outlined in Algorithm~\ref{alg:ldpmabofP} on the auxiliary data before interacting with the target data.
Unlike the target data, the auxiliary datasets already contain executed policies $\pi^{\mathrm{Q}_m}(X_{T_m(t)}^{\mathrm{Q}_m})$.
As a result, learning on the auxiliary data does not involve making instant decisions based on the learned policy $\pi$. However, the active partition $\mathcal{B}_t$ and the associated active arms sets are gradually updated throughout the interaction with auxiliary data.

Importantly, since multiple datasets are involved, Algorithm~\ref{alg:ldpmab} requires a multiple-source version of the local estimator and confidence bound for the reward function. In particular, it is likely that several datasets may contribute to the local estimates of the same bin. Thus, to achieve optimal estimation efficiency, their contributions need to be carefully \textit{weighted} due to different variance levels induced by the LDP constraints.
To this end, we propose a novel multiple-source local estimator where 
\begin{align}\label{equ:privateestimatormulti}
\tilde{f}_{k}^{t} (x)
= 
\sum_{B_s^j \in \mathcal{B}_t}
\eins(x \in B^j_s  ) 
\frac{ \sum_{m = 0}^M 
\lambda_{t,k,s}^{m,j}\sum_{i=1}^{T_m(t)}
\tilde{V}_{i,k,s}^{\mathrm{Q}_m,j} a_{i,s}^{m,j}}
{ \sum_{m = 0}^M 
\lambda_{t,k,s}^{m,j}\sum_{i=1}^{T_m(t)}
\tilde{U}_{i,k,s}^{\mathrm{Q}_m,j} a_{i,s}^{m,j}}.
\end{align}

In \eqref{equ:privateestimatormulti}, the influence of each dataset is controlled by the weight $\lambda_{t,k,s}^{m,j}$. 
Specifically, we set 
\begin{align}\label{equ:lamdaratios}
\lambda_{t,k,s}^{m,j} = \left|\frac{\varepsilon_m^2}{t_s^{m,j}}\sum_{i=1}^{T_m(t)}
\tilde{U}_{i,k,s}^{\mathrm{Q}_m,j} a_{i,s}^{m,j} \right| \wedge \eins\left\{t_s^{m,j} \geq  \log^2 (n)\right\},
\end{align} 
where recall we denote $n=n_{\mathrm{P}}\vee (\max_{m=1}^M n_{\mathrm{Q}_m}).$ Here, the condition $\eins\{t_s^{m,j} \geq  \log^2 (n)\}$ ensures that the $m$-th dataset has provided sufficient samples, a requirement needed for the theoretical validity of our confidence bound in \eqref{equ:rkj}.
When the condition is unmet, the weight is zero, and the $m$-th data is excluded from $\tilde{f}^t_k(x)$. 
When the condition holds, the weight $\lambda_{t,k,s}^{m,j}$ depends on two factors that characterize the information from the $m$-th dataset. 
One is $({t_s^{m,j}})^{-1}\sum_{i=1}^{T_m(t)} \tilde{U}_{i,k,s}^{\mathrm{Q}_m,j} a_{i,s}^{m,j}$, which approximates the proportion of samples within the bin that pulled arm $k$ and represents the quantity of information. 
The other is the privacy budget $\varepsilon_m$, which reflects the accuracy of each $\tilde{U}_{i,k, s}^{ \mathrm{Q}_m, j}$ and represents the quality of information. 
If both factors are relatively large, the dataset is considered informative and is therefore assigned a large weight. We note that without LDP constraints, such weighting scheme is not necessary. Indeed, in the non-private case (i.e.~$\varepsilon_m = \infty$), our choice of $\lambda$ indicates that all weights are assigned equal values of $1$, which is consistent with non-private transfer learning for nonparametric contextual MAB~\citep{cai2024transfer}.

Moreover, we define the corresponding confidence bound as
\begin{align}\label{equ:rkj}
r_{k,s}^{t,j} :=\sqrt{\frac{C_{ n} \sum_{m=0}^M(\lambda_{t,k,s}^{m,j})^2 \left\{
\left(\varepsilon_m^{-2} t_{s}^{m,j}\right) \vee \sum_{i=1}^{T_m(t)}
\tilde{U}_{i,k,s}^{\mathrm{Q}_m,j} a_{i,s}^{m, j}\right\} }{\left(\sum_{m = 0}^M 
\lambda_{t,k,s}^{m,j}
\sum_{i=1}^{T_m(t)}
\tilde{U}_{i,k,s}^{\mathrm{Q}_m,j} a_{i,s}^{m, j}\right)^2}},
\end{align}
where $C_{n}\asymp \log (n) $ with its exact expression specified in the proof.  As shown in \Cref{lem:finitedifferenceboundauxilary}, \eqref{equ:rkj} provides a valid high-probability confidence bound for the multiple-source estimator, with a rationale similar to that of \eqref{equ:rkjofP}. 
Note that the term $\sum_{i=1}^{T_m(t)} \tilde{U}_{i,k,s}^{\mathrm{Q}_m,j} a_{i,s}^{m, j}$ approximately corresponds to the number of samples in the $m$-th dataset falling in $B_s^j$ while pulling arm $k$. This quantity generally increases with~$\kappa_m$ and decreases with $\gamma_m$, in view of the definitions of these quantities. 
Therefore, as a statistic, \eqref{equ:rkj} naturally encodes information about $\kappa_m$ and $\gamma_m$, which is the key reason that enables our estimator and thus algorithm to be adaptive to these \textit{unknown} parameters.

Given the newly designed local estimator \eqref{equ:privateestimatormulti} and the confidence bound \eqref{equ:rkj}, the algorithm can then conduct arm elimination and bin refining. In particular, an arm $k^*$ is removed from the active arm set $A_s^j$ of a bin $B_s^j \in \mathcal{B}_t$ if there exists $k\neq k^*$ such that
\begin{align}
r_{k,s}^{t,j}, r_{k^*,s}^{t,j} > 0 \;\; \text{ and } \;\; \tilde{f}_{k}^{t}(x) -  2r_{k,s}^{t,j}  > \tilde{f}_{k^*}^{t}(x)  + 2r_{k^*,s}^{t,j}. 
\label{equ:updaterule}
\end{align}
Similar to \eqref{equ:updateruleP}, the first condition in \eqref{equ:updaterule} aims to ensure that sufficient samples have been collected, since we notice $r_{k,s}^{t,j} > 0$ implies at least one dataset provides $\log ^2 (n)$ samples. A bin $B_s^j \in \mathcal{B}_t$ is refined if $r_{k,s}^{t, j}< {\tau}_{s}$ for some $k\in A_{s}^j$, where the parameter ${\tau}_{s}$ is set as in \eqref{equ:choiceoftaus}.

\begin{algorithm}[!t]
\caption{The nonparametric MAB algorithm under LDP with auxiliary data \\(For simplicity, we do not explicitly separate the user and server sides in the presentation.)}
\label{alg:ldpmab}
{\bfseries Input:}{Budgets $\varepsilon, \varepsilon_1,\ldots,\varepsilon_M$, auxiliary sample sizes $n_{\mathrm{Q}_1}, \ldots, n_{\mathrm{Q}_m}$, target sample size $n_{\mathrm{P}}$.}
\\
{\bfseries Initialization: }{  $\pi_1 = \operatorname{Unif}([K])$, $\mathcal{B}_1 =\{B_0^1\}=\{ [0,1]^d\}$, $A_0^1 = [K]$, $t=1$.}\\
{\color{red} \texttt{\# jump-start via auxiliary data}}\\
\For{$m\in[M]$}{
\For{$i \in [n_{\mathrm{Q}_m}]$}{
\For{$B_s^j\in \mathcal{B}_t$}{
Compute \eqref{equ:privateestimatormulti} and \eqref{equ:rkj}.
\hfill {\color{blue} \texttt{\# estimating reward functions}}\\
Remove $k$ from $A_s^j$ if \eqref{equ:updaterule} holds.\hfill {\color{blue} \texttt{\# eliminating arms}} \\
\If{ $r_{k,s}^{t, j}< {\tau}_{s}$ \textup{for some} $k\in A_{s}^j$
}{
$\mathcal{B}_t = \mathcal{B}_t \cup \{B_{s+1}^{2j-1}, B_{s+1}^{2j}\} \setminus B_{s}^{j}$. \hfill {\color{blue} \texttt{\# refining bins}}\\
$A_{s+1}^{2j-1} = A_s^{j}$, $A_{s+1}^{2j} = A_s^{j}$. 
}
}

Set $t \leftarrow t + 1$, $\mathcal{B}_t = \mathcal{B}_{t-1}$ and update $\pi_t$ by \eqref{equ:partitionbasedpolicy}.
}
}
{\color{red} \texttt{\# interaction on target data}}\\
\For{$i\in [n_{\mathrm{P}}]$}{
The user $i$ receives $\pi_t$ from the server, pulls an arm via $\pi_t$ and receives the reward. \\
\For{$B_s^j\in \mathcal{B}_t$}{
Compute \eqref{equ:privateestimatormulti} and \eqref{equ:rkj}.
\hfill {\color{blue} \texttt{\# estimating reward functions}}\\
Remove $k$ from $A_s^j$ if \eqref{equ:updaterule} holds.
\hfill {\color{blue} \texttt{\# eliminating arms}} \\
\If{ $r_{k,s}^{t, j}< {\tau}_{s}$ \textup{for some} $k\in A_{s}^j$
}{
$\mathcal{B}_t = \mathcal{B}_t \cup \{B_{s+1}^{2j-1}, B_{s+1}^{2j}\} \setminus B_{s}^{j}$. \hfill {\color{blue} \texttt{\# refining bins}}\\
$A_{s+1}^{2j-1} = A_s^{j}$, $A_{s+1}^{2j} = A_s^{j}$. 
}
}
Set $t \leftarrow t + 1$ and $\mathcal{B}_t = \mathcal{B}_{t-1}$.
Update $\pi_t$ by \eqref{equ:partitionbasedpolicy} and send to the next user. 
}
\end{algorithm}

\section{Numerical experiments}\label{sec:experiments}

In this section, we conduct numerical experiments on both synthetic data (Section \ref{sec:syntheticexperiments}) and real-world data (Section \ref{sec:realexperiments}), to respectively validate our theoretical findings and show promising performance of the proposed method. All experiments are conducted on a machine with 72-core Intel Xeon 2.60GHz and 128GB memory.
Reproducible codes are available on GitHub\footnote{https://github.com/Karlmyh/LDP-Contextual-MAB}.

\subsection{Simulation Studies}\label{sec:syntheticexperiments}

\begin{figure}[htbp]
\centering
\vskip -0.1in
\subfigure[$k=1$]{
\begin{minipage}{0.3\linewidth}
\centering
\includegraphics[width=\textwidth]{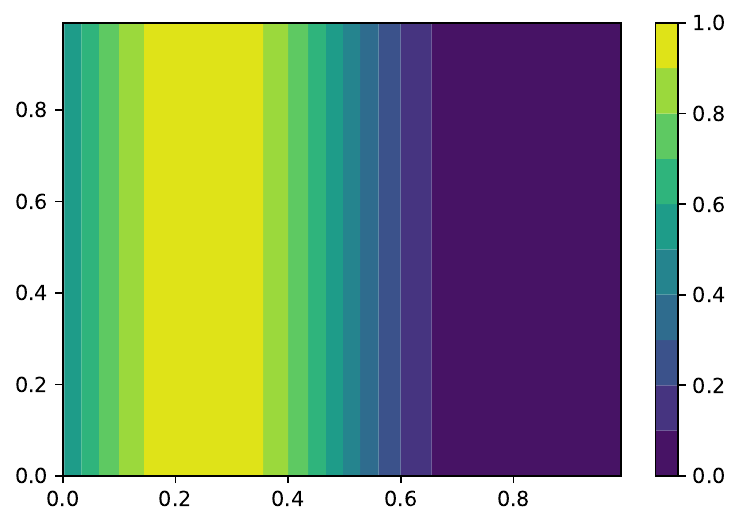}
\end{minipage}
\label{fig:rewardfunction1}
}
\subfigure[$k=2$]{
\begin{minipage}{0.3\linewidth}
\centering
\includegraphics[width=\textwidth]{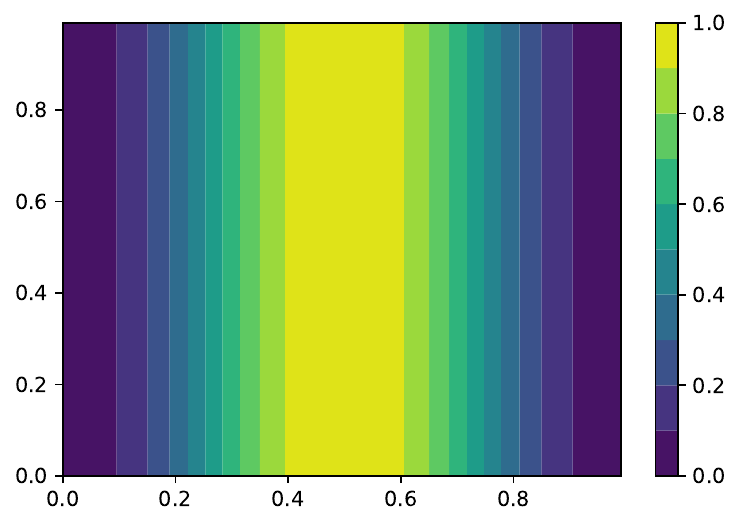}
\end{minipage}
\label{fig:rewardfunction2}
}
\subfigure[$k=3$]{
\begin{minipage}{0.3\linewidth}
\centering
\includegraphics[width=\textwidth]{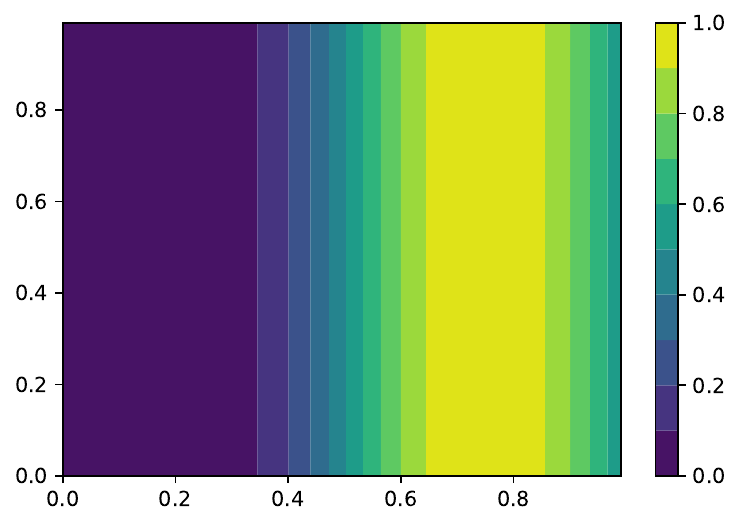}
\end{minipage}
\label{fig:rewardfunction3}
}
\caption{Illustration of reward functions. } 
\label{fig:rewardfunction}
\end{figure}

\begin{figure}[htbp]
\centering
\subfigure[$\gamma = 0.2$]{
\begin{minipage}{0.3\linewidth}
\centering
\includegraphics[width=\textwidth]{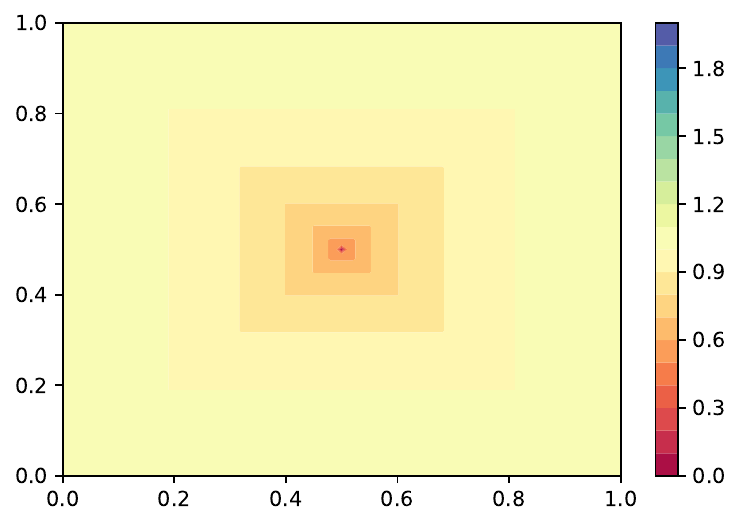}
\end{minipage}
\label{fig:marginal_distribution_1}
}
\subfigure[$\gamma = 1$]{
\begin{minipage}{0.3\linewidth}
\centering
\includegraphics[width=\textwidth]{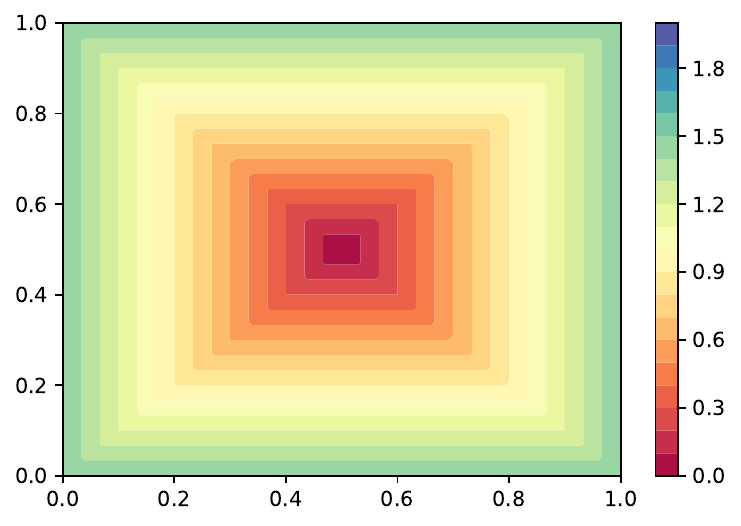}
\end{minipage}
\label{fig:marginal_distribution_2}
}
\subfigure[$\gamma = 2$]{
\begin{minipage}{0.3\linewidth}
\centering
\includegraphics[width=\textwidth]{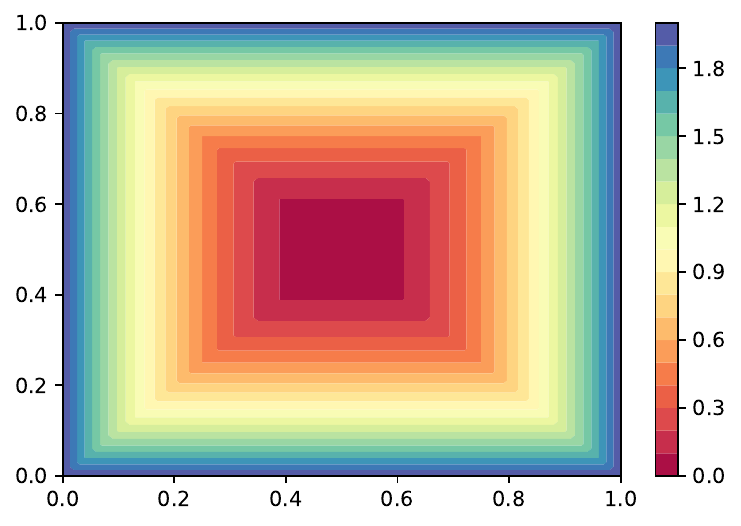}
\end{minipage}
\label{fig:marginal_distribution_3}
}
\caption{Illustration of marginal distribution $Q_{m,X}$ of source data. } 
\label{fig:marginal_distribution}
\end{figure}

\noindent\textbf{Synthetic Distributions}. For distribution $\mathrm{P}$, we choose the marginal distribution $\mathrm{P}_X$ to be the uniform distribution on $\mathcal{X} = [0,1]^d$. 
For the reward function, let 
\begin{align*}
f_k(x) = \frac{2\exp( - 2 K^2  ( x^1 - k / K)^2   ) }{ 1 + \exp( - 2 K^2  ( x^1 - k / K)^2   )}. 
\end{align*}
The reward functions are plotted in Figure \ref{fig:rewardfunction}.
The auxiliary data distribution is taken as $\mathrm{Q}_{m,X}(x) = c_{norm} \|x - \mathrm{I}_d/ 2 \|_{\infty}^{\gamma}$, where $\mathrm{I}_d$ is the $d$ dimensional vector with all entries equal to $1$.
We can explicitly compute the normalizing constant $c_{norm} =  2^{- \gamma}d / (d + \gamma) $. 
Figure \ref{fig:marginal_distribution} illustrates $\gamma = 0.2, 1, 2$.
The behavior policies for the auxiliary data are a discrete distribution with probability vector $\kappa/K + (2-2\kappa)/\{K(K-1)\} \cdot (0, \ldots, K-1)$ over $[K]$, which belongs to $\Lambda(K,\beta,\gamma,\kappa )$. 

In the numerical experiments, we fix $K=3$ and take $\varepsilon, \varepsilon_m \in \{1,2,4,8,1024\}$, covering commonly seen magnitudes of privacy budgets from high to low privacy regimes \citep{erlingsson2014rappor, apple2017differential} as well as the (essentially) non-private case. To conserve space, the implementation details of all algorithms can be found in \Cref{sec:implementationdetail} of the supplement. 
In \Cref{sec:add_exp} of the supplement, we further provide numerical results under an alternative simulation setting with more complex reward functions, where similar findings as the ones seen below in Figures \ref{fig:samplesizen}-\ref{fig:underlyingparameter} are observed. All simulation results presented below are based on 100 repetitions unless otherwise noted. 

\noindent\textbf{An Illustrative Example}. We first illustrate how the auxiliary datasets benefit the learning process via a simple example.
For $n_{\mathrm{P}}$ target samples, we consider the following metrics for $t\in [n_{\mathrm{P}}]$.
For global performance, we use the overall averaged regret
\begin{align*}
\overline{R}_{t}^{\mathrm{global}}(\pi) = \frac{1}{t}\sum_{i=1}^{t}  
\left( f_{\pi^{*}(X_i^{\mathrm{P}})}(X_i^{\mathrm{P}})-f_{\pi_{i}(X_i^{\mathrm{P}})}(X_i^{\mathrm{P}}) \right).
\end{align*}
For local performance, we use two metrics at a fixed point $x\in \mathcal{X}$, the local averaged regret and the ratio of chosen arms:
\begin{align*}
\overline{R}_{t}^{\mathrm{local}}(\pi, x) = 
\frac{1}{t}\sum_{i=1}^{t}  
\left( f_{\pi^{*}(x)}(x)-f_{\pi_{i}(x)}(x) \right) , \;\; \;\; \overline{R}_{t}^{\mathrm{ratio}}(\pi, x,  k) =   \frac{1}{t}\sum_{i=1}^{t} \eins  \left( \pi_i(x) = k\right).  
\end{align*}

\begin{figure}[!t]
\centering
\subfigure[Learning process without auxiliary data.]{
\begin{minipage}{0.3\linewidth}
\centering
\includegraphics[width=\textwidth]{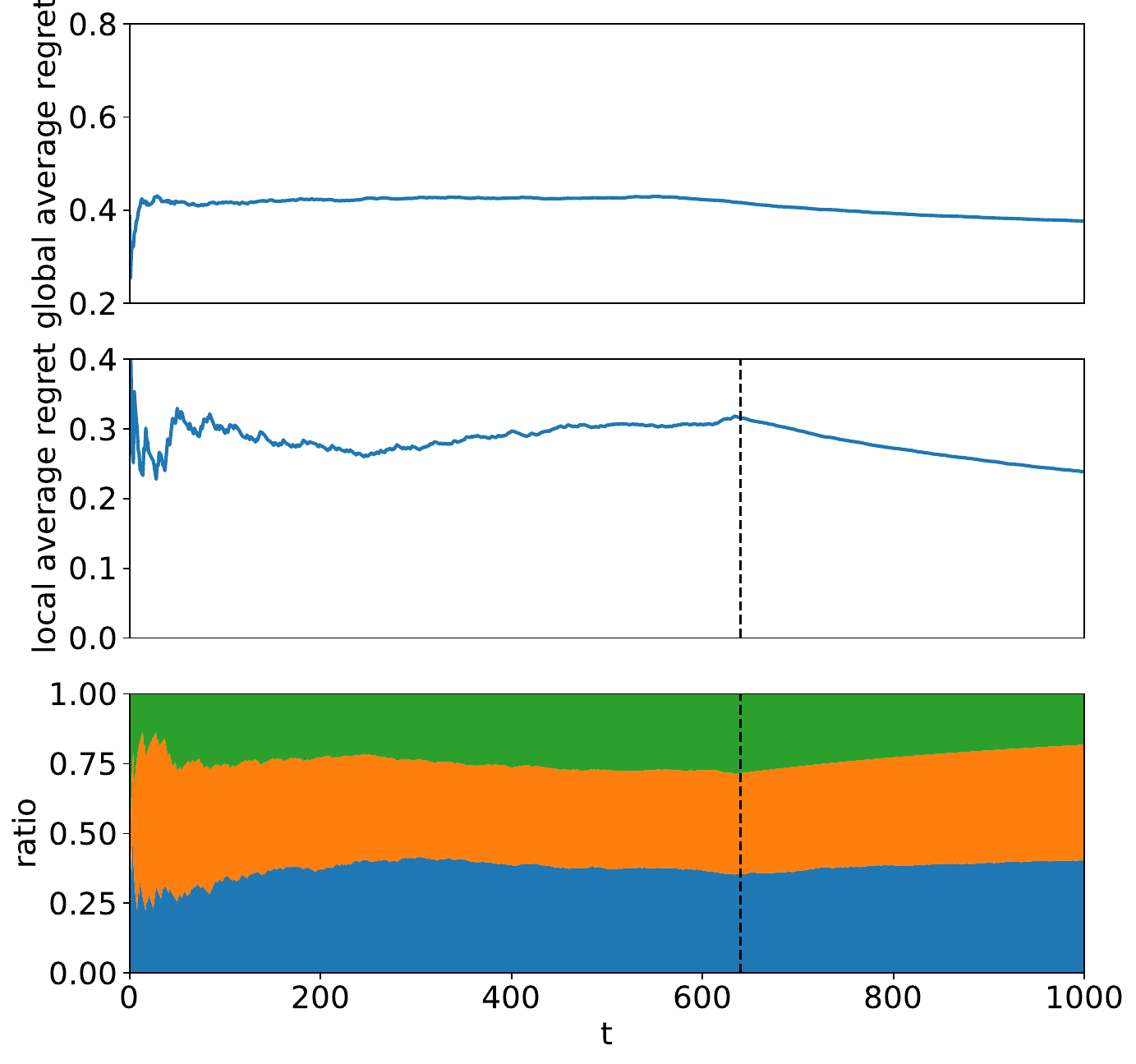}
\end{minipage}
\label{fig:bandit_process_1}
}
\subfigure[Learning process with effective auxiliary data.]{
\begin{minipage}{0.3\linewidth}
\centering
\includegraphics[width=\textwidth]{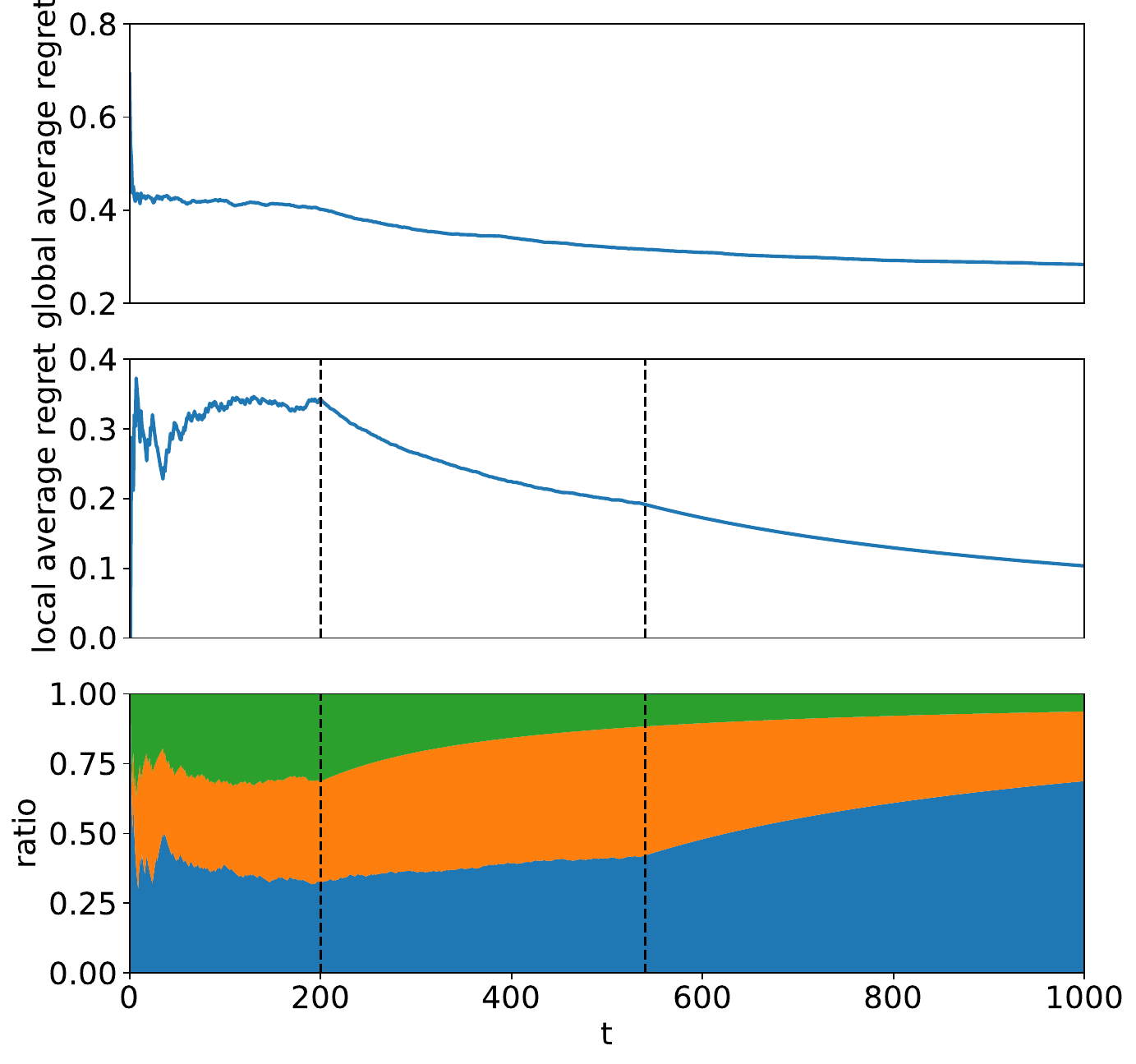}
\end{minipage}
\label{fig:bandit_process_2}
}
\subfigure[Learning process with weak auxiliary data.]{
\begin{minipage}{0.3\linewidth}
\centering
\includegraphics[width=\textwidth]{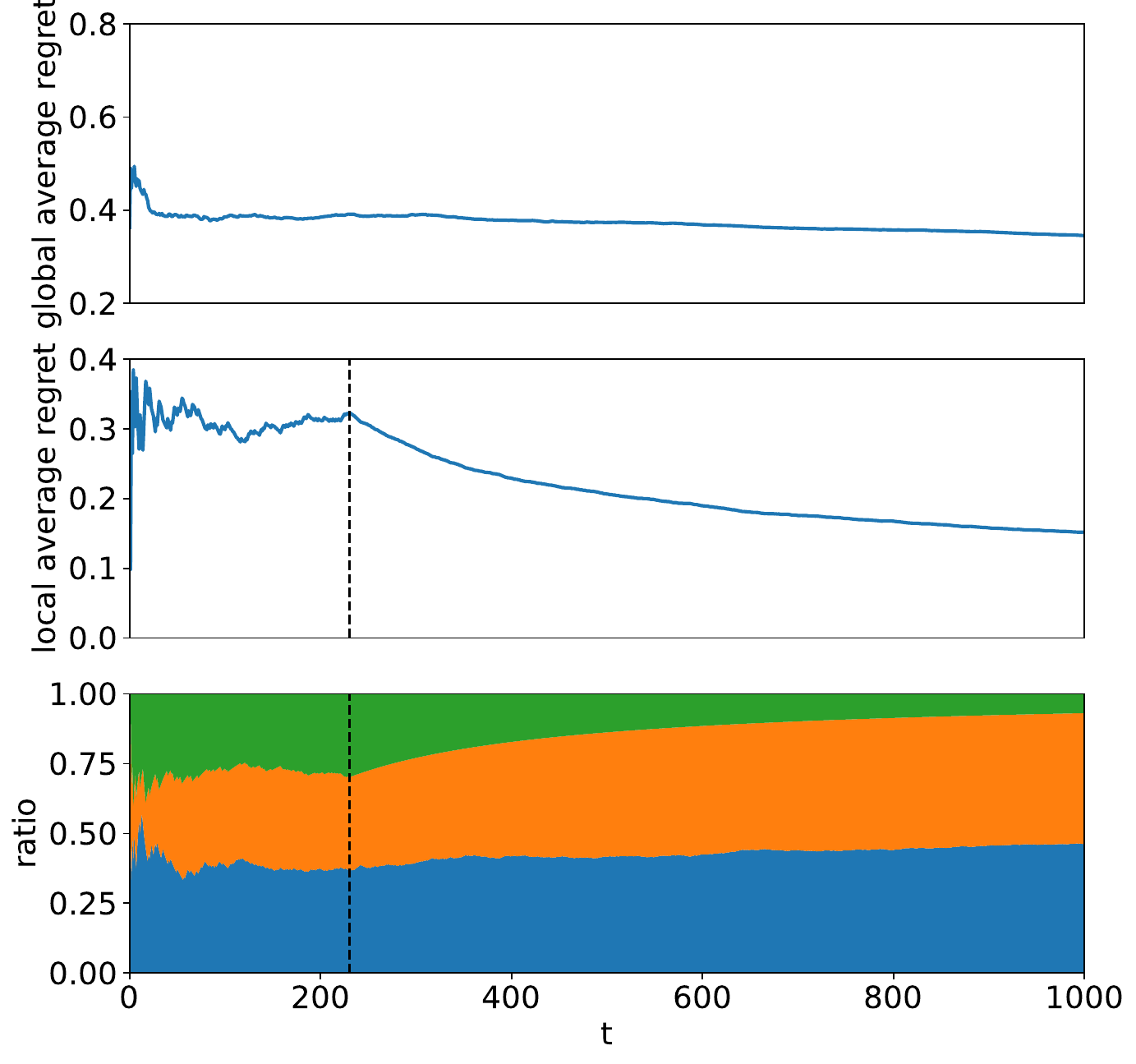}
\end{minipage}
\label{fig:bandit_process_3}
}
\caption{ 
We set
$\varepsilon = 1$, $n_{\mathrm{P}} = 1000$, and $M = 1$.
The effective auxiliary data has $n_{\mathrm{Q}_1} = 500$, $\varepsilon_1 = 8$, and $\gamma_1 = 0$.
The weak auxiliary data has $n_{\mathrm{Q}_1} = 500$, $\varepsilon_1 = 0.5$, and $\gamma_1 = 5$. 
Both auxiliary dataset has $\kappa_1 =1$. 
We run a single trial as a showcase. 
The top row exhibits the global average regret curves. 
The middle row exhibits the local average regret curve at $x = (1/3, 1/3)$.
The bottom row exhibits the ratio of pulled arms at $x=  (1/3, 1/3)$, which is represented by the width of each color at the cross-section at the time $t$. 
Blue, orange, and green represent the arm 1, 2, and 3, respectively. 
Note that we know the best arm for $(1/3, 1/3)$ is $1$, i.e., we expect to see the blue area increase.
The black vertical lines indicate when one of the sub-optimal arms at $(1/3, 1/3)$ is eliminated, leading to a phase transition in the local regret curves and arm ratios.
It is observed that both types of auxiliary data bring forward the elimination of sub-optimal arms~(such an event is marked by vertical dashed line), but the effective auxiliary data is significantly more impactful. 
} 
\label{fig:bandit_process}
\end{figure}

For a naive policy that selects arms uniformly at random, all three quantities should remain approximately unchanged for all time steps.
For any effective policy, we expect to see $\overline{R}_{t}^{\mathrm{global}}(\pi)$ and $\overline{R}_{t}^{\mathrm{local}}(\pi, x)$ decreasing and $\overline{R}_{t}^{\mathrm{ratio}}(\pi, x,  \pi^*(x))$ increasing over time. 
We use the average metrics instead of cumulative regret as the zero-order trend is more apparent than the first-order trend for visualization.
We consider three settings: learning without auxiliary data, with effective auxiliary data, and with weak auxiliary data.
The results in Figure \ref{fig:bandit_process} show that auxiliary data significantly accelerates the learning process by eliminating sub-optimal arms in the early stages, effectively providing a jump-start that leads to faster descent in both local and global regret.
Additionally, the quality of the auxiliary data determines the magnitude of this jump-start effect.

\begin{figure}[htbp]
\centering
\subfigure[Regret curve]{
\begin{minipage}{0.35\linewidth}
\centering
\includegraphics[width=\textwidth]{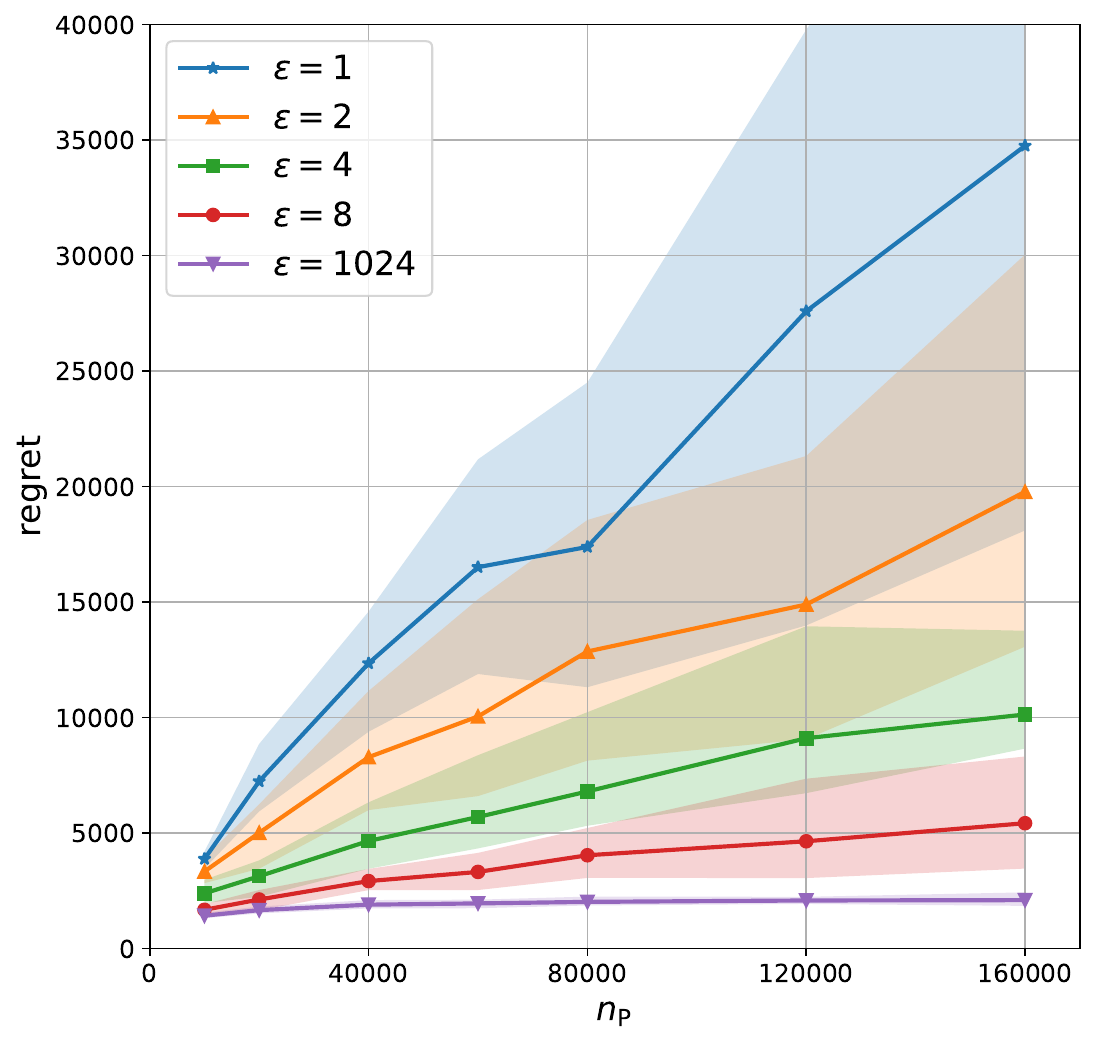}
\end{minipage}
\label{fig:samplesizen1}
}
\hskip +0.2in
\subfigure[Regret curve with auxiliary data]{
\begin{minipage}{0.35\linewidth}
\centering
\includegraphics[width=\textwidth]{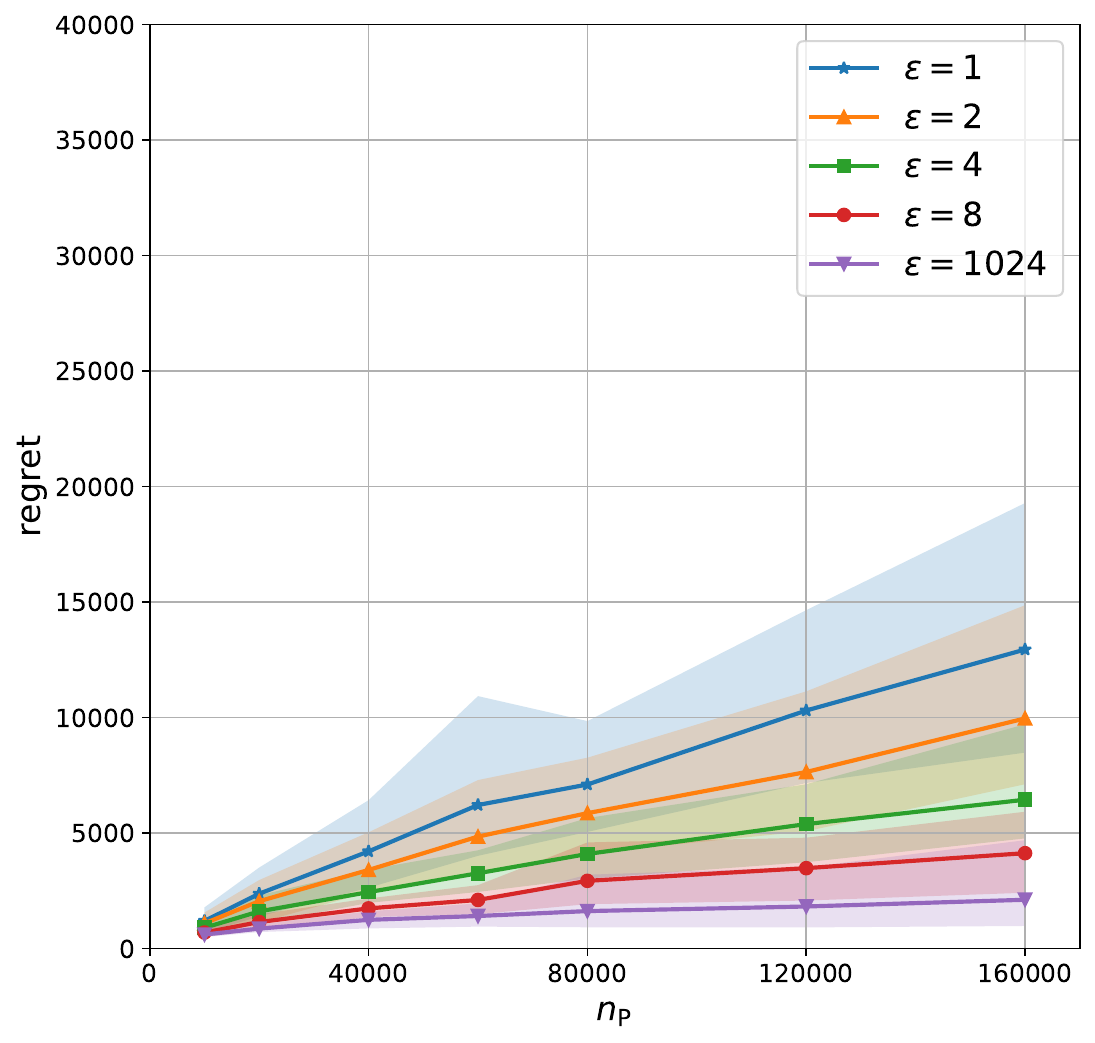}
\end{minipage}
\label{fig:samplesizen2}
}
\caption{Regret with $\varepsilon\in\{1,2,4,8, 1024\}$ and $n_{\mathrm{P}}\in\{1,2,4,6,8,12,16\}\times 10^4$. 
In (b), we use auxiliary data with $n_{\mathrm{Q}_1} = 5000$, $\varepsilon_1 = 8$, $\gamma_1 = 0$ and $\kappa_1 = 1$. The colored areas are 95\% confidence intervals.}
\label{fig:samplesizen}
\end{figure}

\noindent\textbf{Sample Sizes}. We first analyze the regret curve with respect to sample sizes $n_{\mathrm{P}}$ in Figure \ref{fig:samplesizen}.
The regret increases in a sub-linear manner with respect to $n_{\mathrm{P}}$, while the growth trend becomes slower as $\varepsilon$ increases.  This aligns with the theoretical finding in Theorem~\ref{thm:upper-bound}. 
Moreover, under the same $\varepsilon$, the growth trend is less steep with the participation of auxiliary data in Figure \ref{fig:samplesizen2}. 
Interestingly, we note that with auxiliary data, the confidence interval of non-private data ($\varepsilon = 1024$) becomes wider since the high variance brought by the (privatized) auxiliary data becomes significant in this case. 
A similar phenomenon is also observed in Figure \ref{fig:sample_size_source}, where we fix the sample size of the target data to examine the improvements brought by auxiliary data under different settings. 
As expected, the improvements are more notable for smaller $\gamma$, larger $n_{\mathrm{Q}_m}$ and $\varepsilon_m$, i.e.~when the auxiliary data has higher quality. 
This phenomenon is well explained by the regret rate characterized in Theorem \ref{thm:upper-bound-auxilary}. 
We also note that confidence intervals are much wider for small $\varepsilon$ and $\varepsilon_m$ in both Figures \ref{fig:samplesizen} and \ref{fig:sample_size_source}, due to the high variance of the injected Laplacian noise.

\begin{figure}[htbp]
\centering
\subfigure[$\gamma = 0, \varepsilon = 1$]{
\begin{minipage}{0.3\linewidth}
\centering
\includegraphics[width=\textwidth]{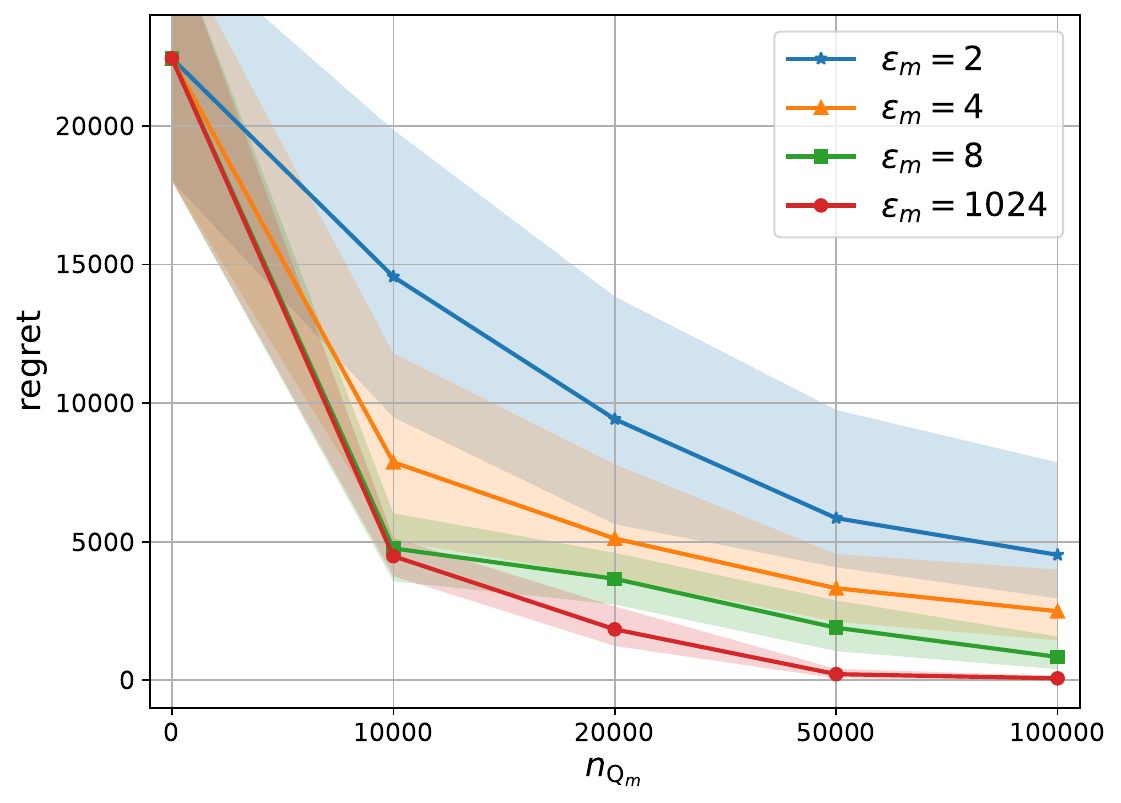}
\end{minipage}
\label{fig:sample_size_source1}
}
\subfigure[$\gamma = 0.2, \varepsilon = 1$]{
\begin{minipage}{0.3\linewidth}
\centering
\includegraphics[width=\textwidth]{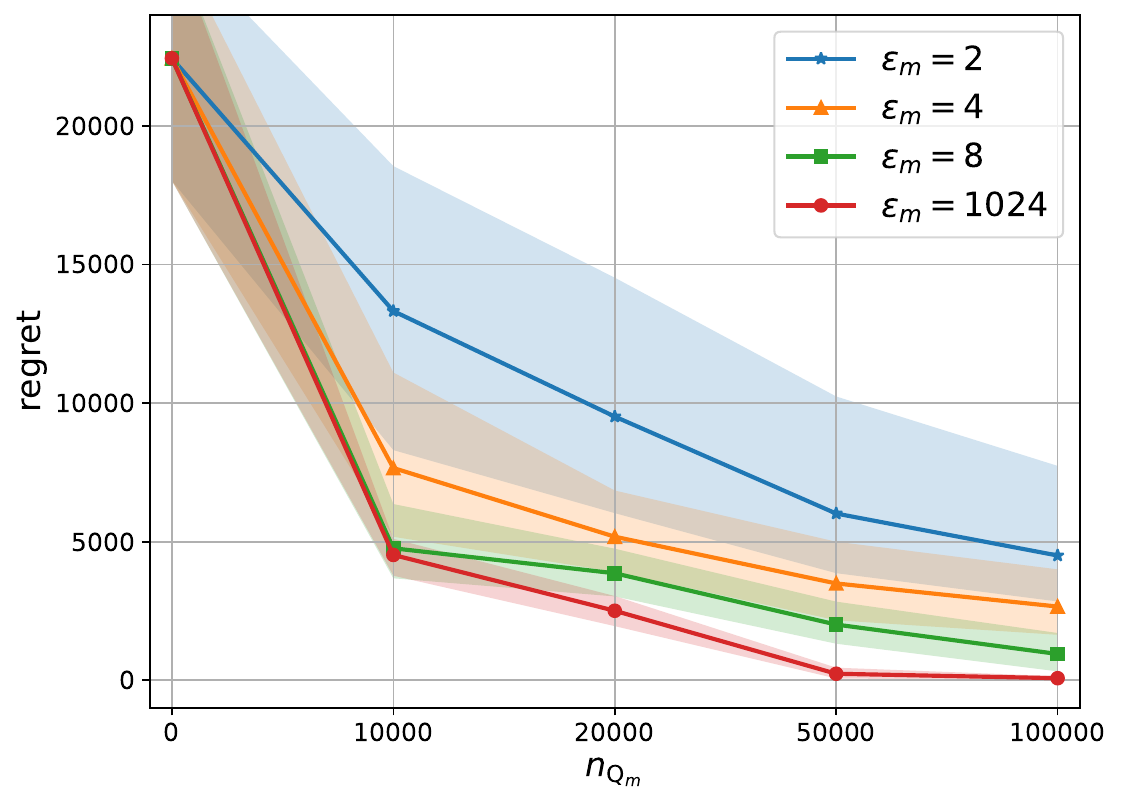}
\end{minipage}
\label{fig:sample_size_source2}
}
\subfigure[$\gamma = 2, \varepsilon = 1$]{
\begin{minipage}{0.3\linewidth}
\centering
\includegraphics[width=\textwidth]{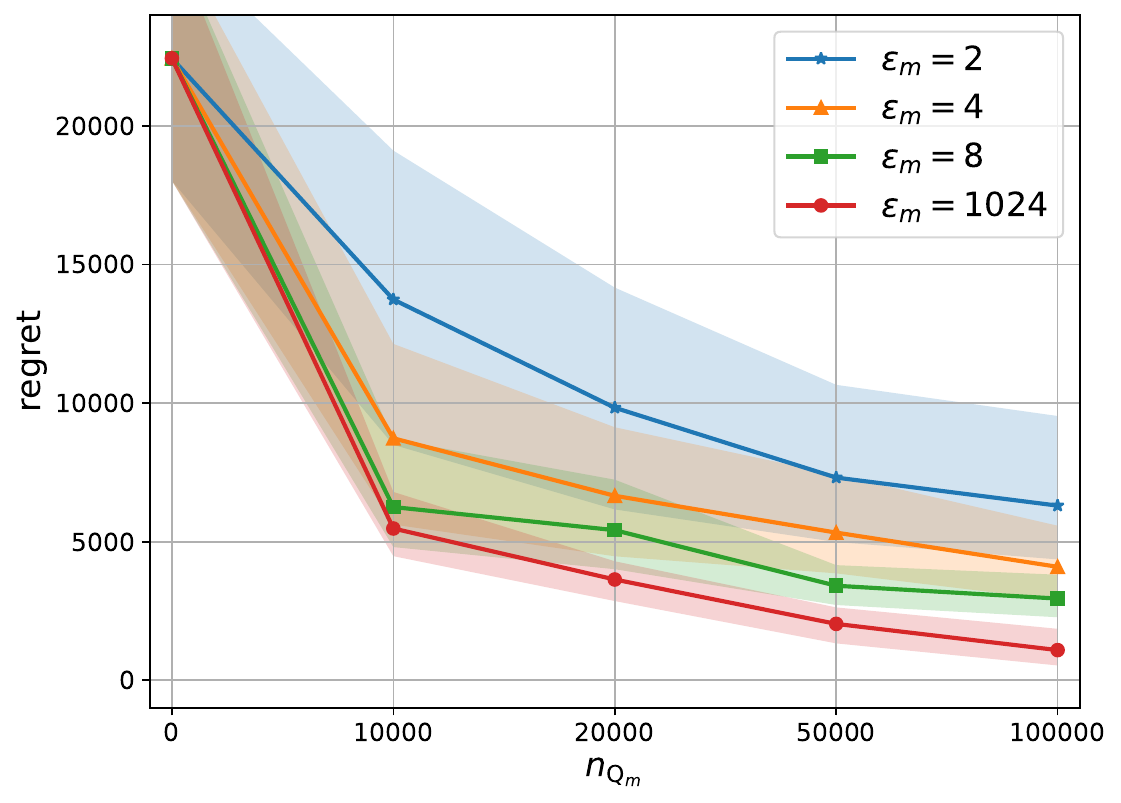}
\end{minipage}
\label{fig:sample_size_source3}
}
\centering
\subfigure[$\gamma = 0, \varepsilon = 2$]{
\begin{minipage}{0.3\linewidth}
\centering
\includegraphics[width=\textwidth]{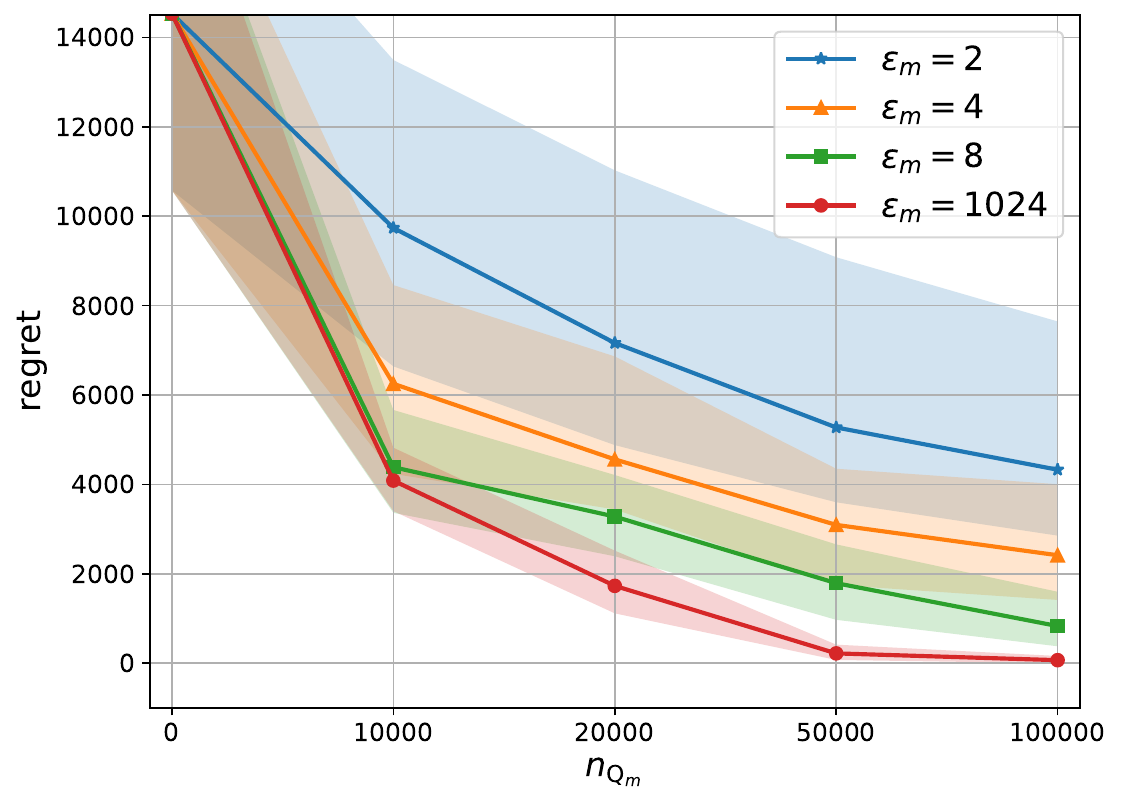}
\end{minipage}
\label{fig:sample_size_source_21}
}
\subfigure[$\gamma = 0.2, \varepsilon = 2$]{
\begin{minipage}{0.3\linewidth}
\centering
\includegraphics[width=\textwidth]{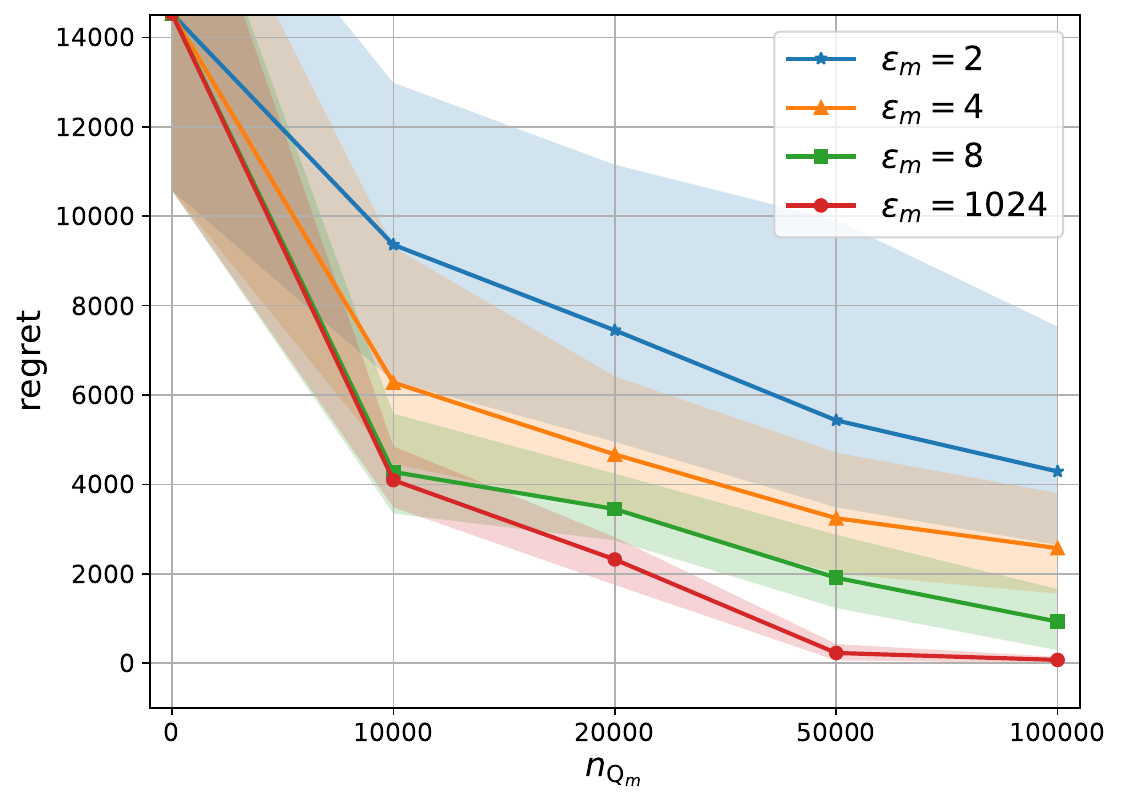}
\end{minipage}
\label{fig:sample_size_source_22}
}
\subfigure[$\gamma = 2, \varepsilon = 2$]{
\begin{minipage}{0.3\linewidth}
\centering
\includegraphics[width=\textwidth]{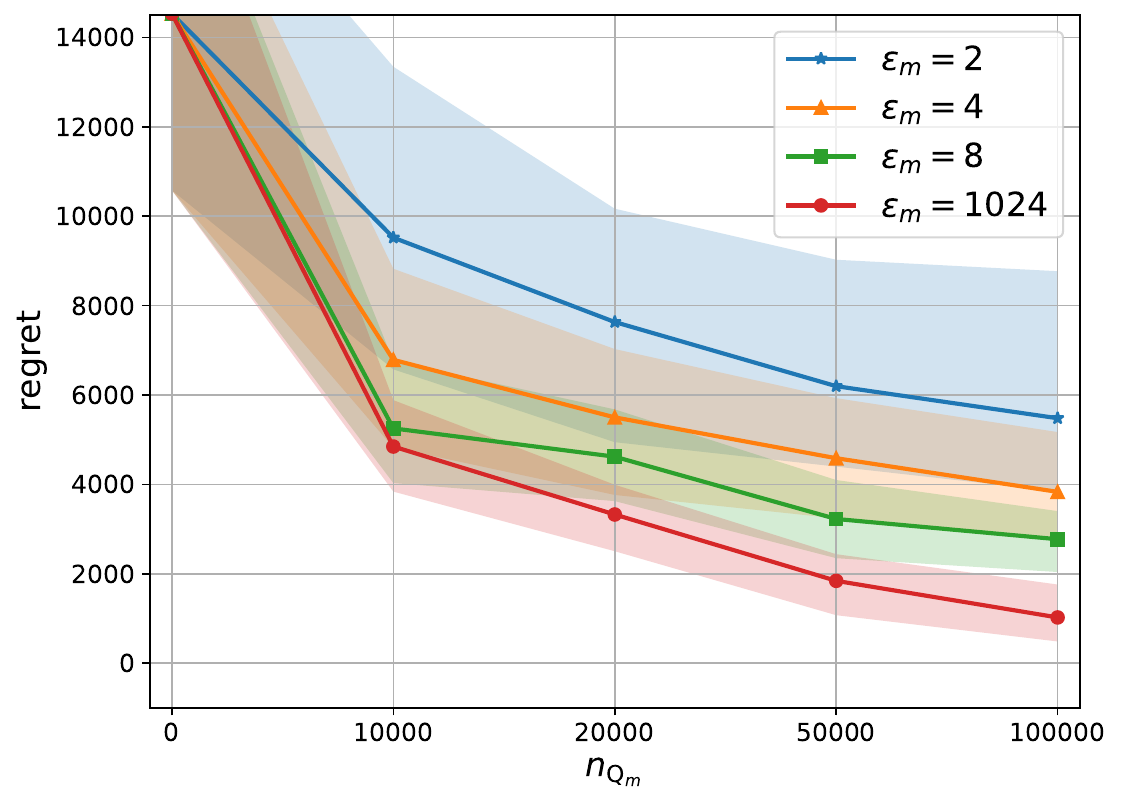}
\end{minipage}
\label{fig:sample_size_source_23}
}
\caption{Regret curves over $n_{\mathrm{Q}_m}\in \{0,1,2,5,10\}\times 10^{4}$ at different $(\gamma,\varepsilon_m)$, while we fix $n_{\mathrm{P}} = 80000$, and fix $M =2$ and $\kappa_1=\kappa_2= 1$. The colored areas are 95\% confidence intervals.} 
\label{fig:sample_size_source}
\end{figure}

\noindent\textbf{Underlying Parameters}. We proceed to investigate the roles of the underlying parameters that control the quality of the auxiliary data, namely $\kappa$ and $\gamma$.
In the bottom panel of Figure \ref{fig:kappa}, we observe that with large $\varepsilon_m$, the regret is notably decreasing with respect to $\kappa$. 
This aligns with the regret upper bound in \eqref{equ:upper-bound-auxilary}. 
In contrast, when $\varepsilon_m$ is small, e.g.~in the top panel of Figure \ref{fig:kappa}, regret barely varies as $\kappa$ changes. 
This is explained by the observation that \eqref{equ:upper-bound-auxilary} is dominated by the target data if $\varepsilon_m$ is too small. 
In this case, the auxiliary dataset does not affect the learning process much, and the variation due to $\kappa$ is negligible. 
For $\gamma$ in Figure \ref{fig:gamma}, we observe a similar phenomenon, where the regret is increasing with respect to $\gamma$, while the slope is controlled by $\varepsilon_m$. 

\begin{figure}[htbp]
\centering
\subfigure[Regret curves over different $\kappa$.]{
\begin{minipage}{0.3\linewidth}
\centering
\includegraphics[width=\textwidth]{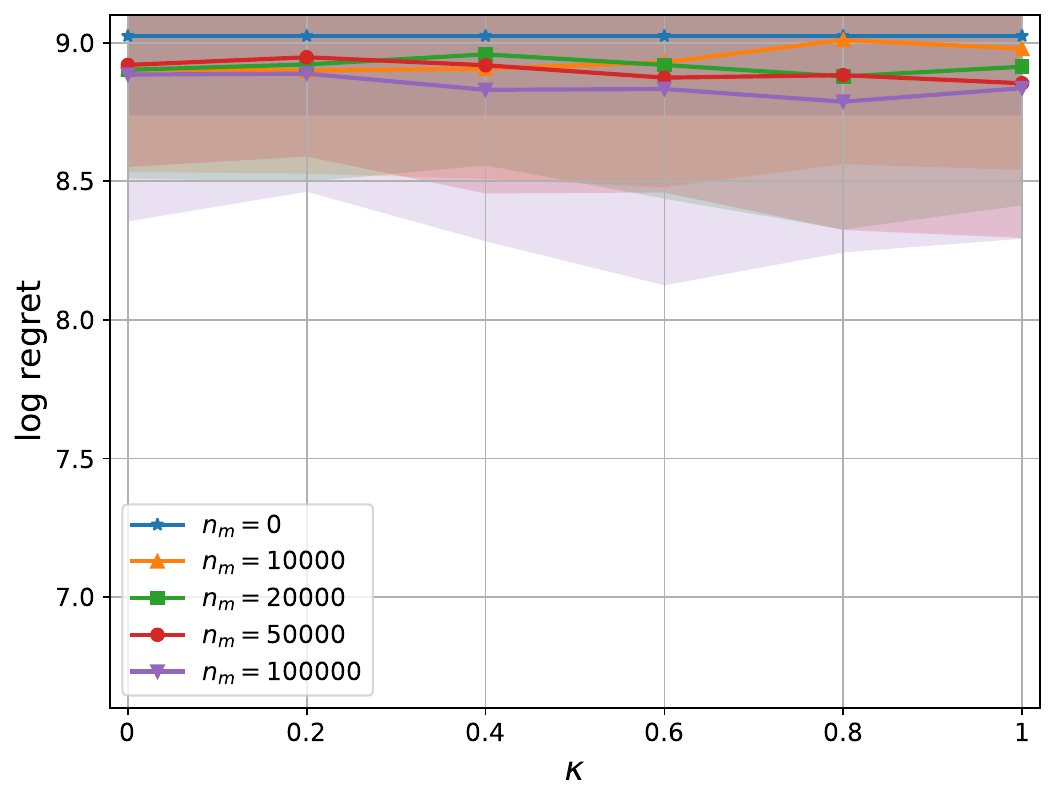}
\includegraphics[width=\textwidth]{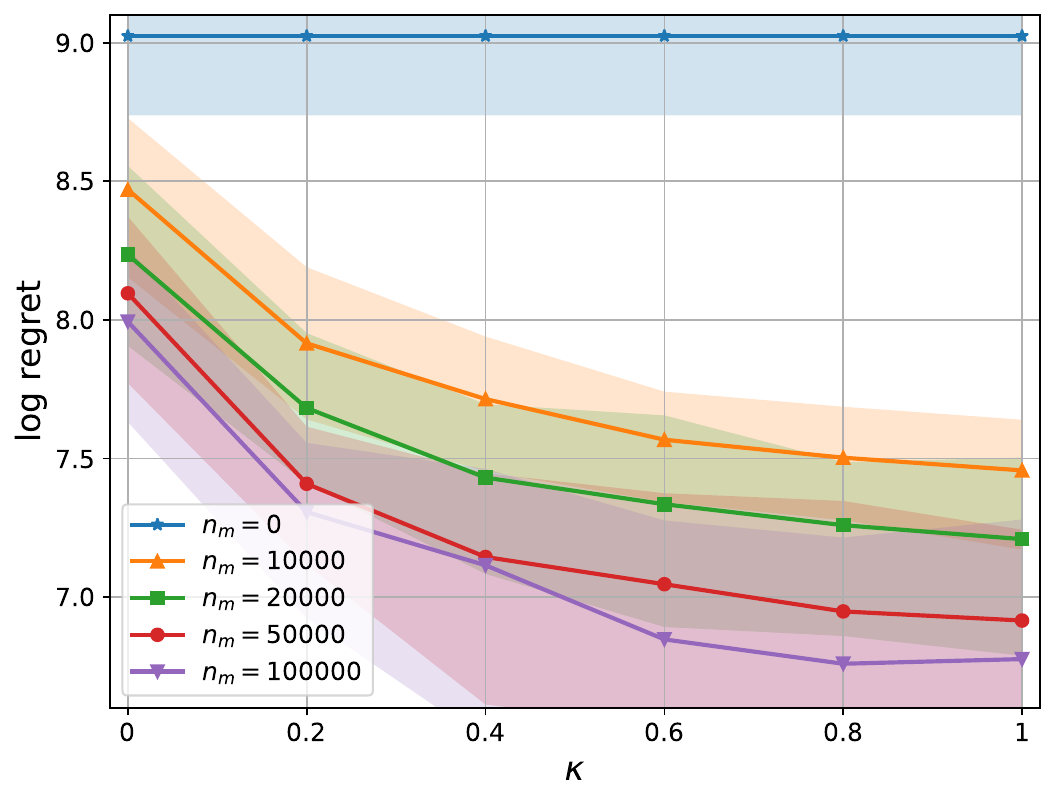}
\end{minipage}
\label{fig:kappa}
}
\subfigure[Regret curves over different $\gamma$.]{
\begin{minipage}{0.3\linewidth}
\centering
\includegraphics[width=\textwidth]{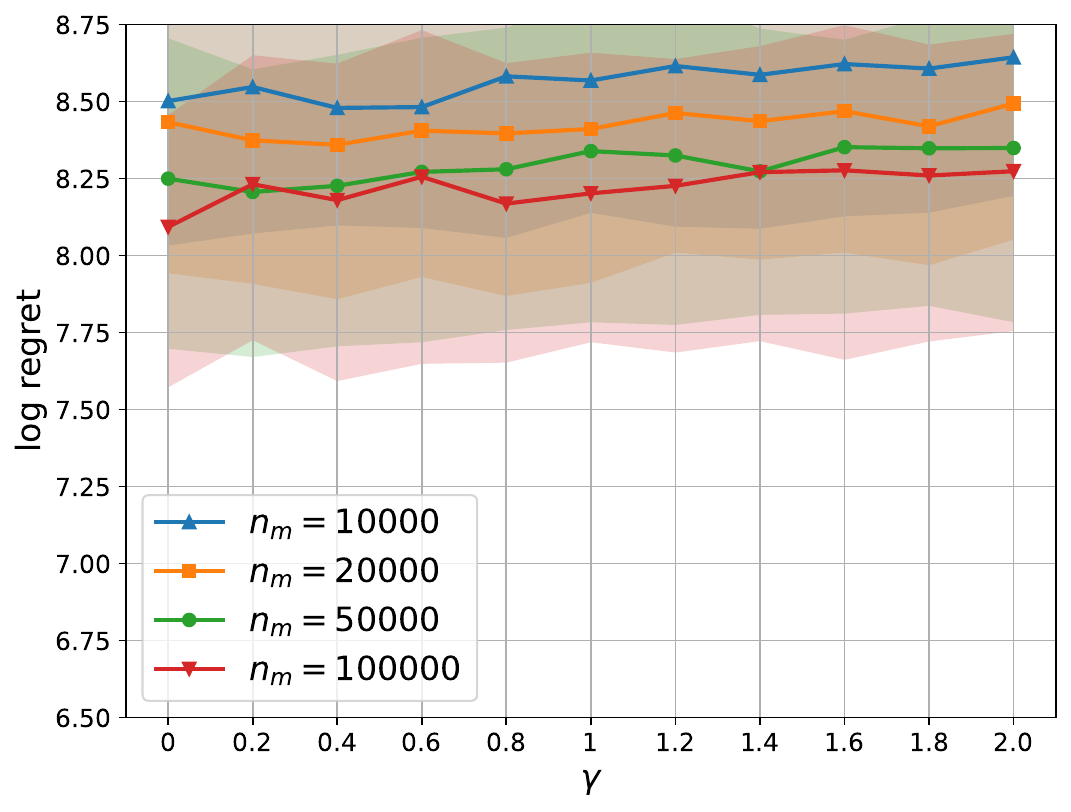}
\includegraphics[width=\textwidth]{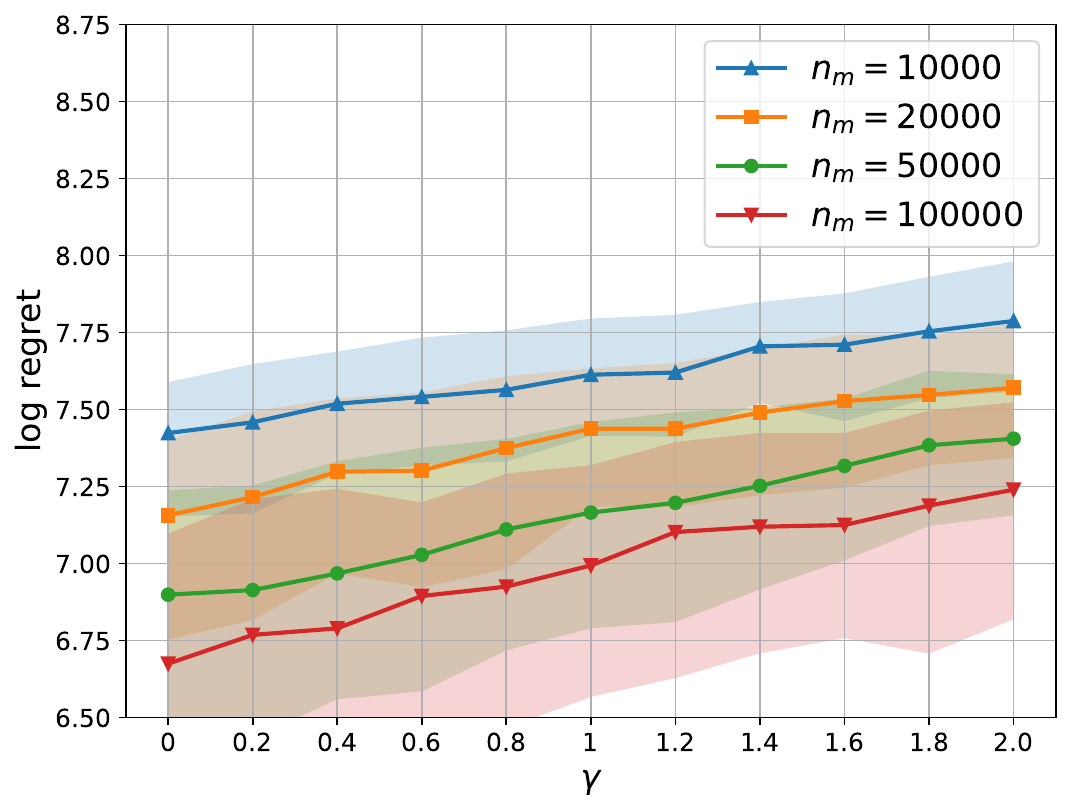}
\end{minipage}
\label{fig:gamma}
}
\subfigure[Order of auxiliary data.]{
\begin{minipage}{0.3\linewidth}
\centering
\includegraphics[width=\textwidth]{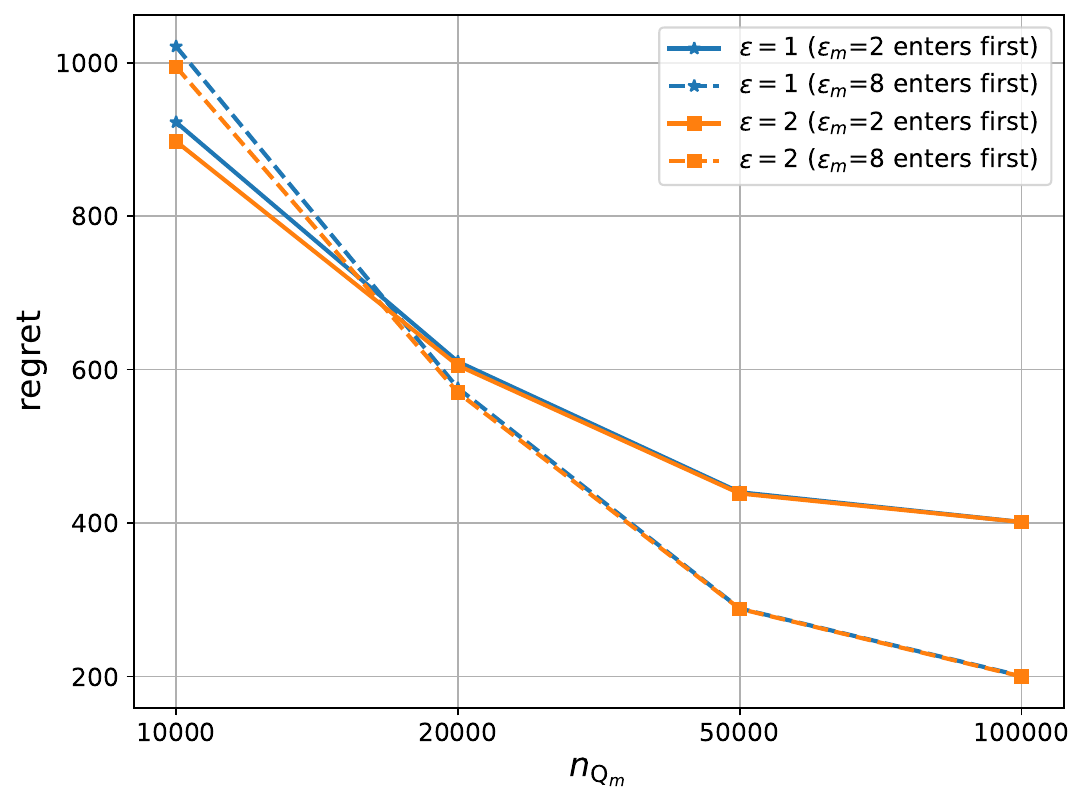}
\includegraphics[width=\textwidth]{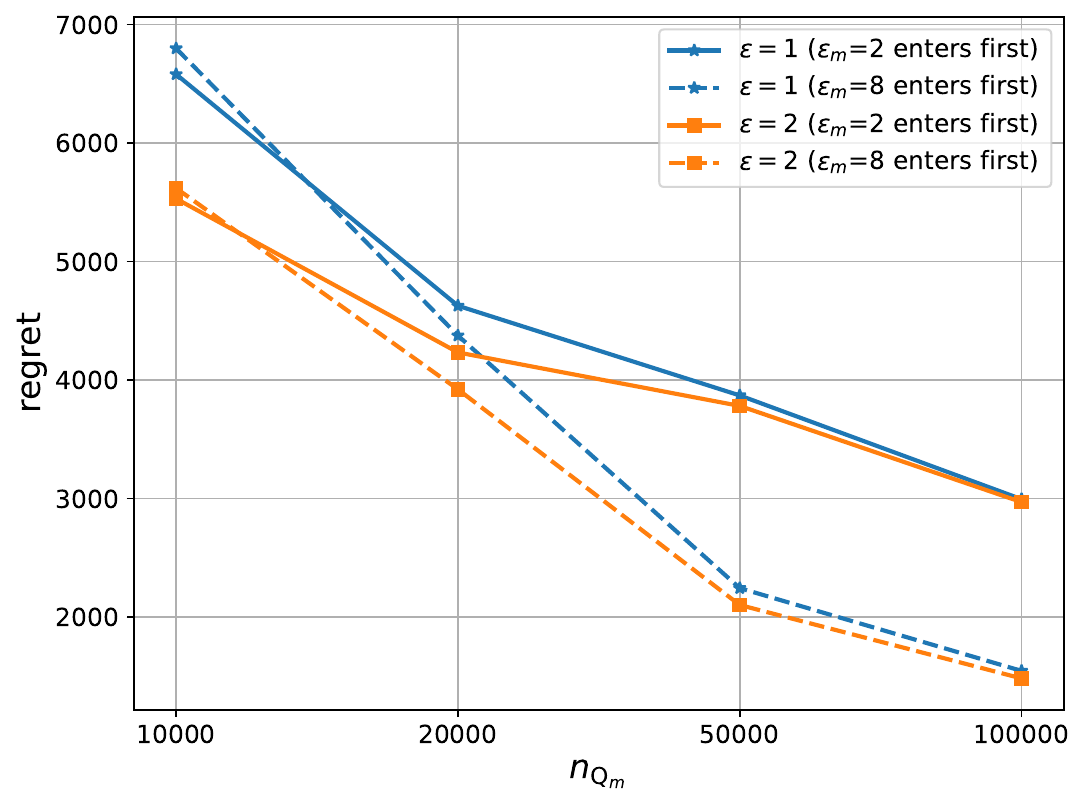}
\end{minipage}
\label{fig:order}
}
\vskip -0.1 in
\caption{(a) Regret curves over $\kappa\in\{0,0.2,\cdots,0.8,1\}$ with different auxiliary privacy budgets (top $\varepsilon_m = 1$, bottom $\varepsilon_m = 8$), while fixing $M =2$, $\gamma = 0.2$, $n_{\mathrm{P}} = 40000$ and $\varepsilon =2$; (b) Regret curves over $\gamma\in \{0,0.2,\cdots,1.8,2\}$ with different auxiliary privacy budgets (top $\varepsilon_m = 2$, bottom $\varepsilon_m = 8$), while fixing $M =2$, $\kappa = 1$, $n_{\mathrm{P}} = 40000$ and $\varepsilon =2$; (c) Comparison of regret curves when the two auxiliary datasets enter the jump-start stage in different orders, for different target data budgets $\varepsilon\in \{1,2\}$ (top $n_{\mathrm{P}} = 10000$, bottom $n_{\mathrm{P}} = 80000$). The colored areas are 95\% confidence intervals.
} 
\label{fig:underlyingparameter}
\end{figure}

\noindent\textbf{Order of Auxiliary Data}. We demonstrate potential improvements by carefully arranging the order in which auxiliary datasets are introduced during the jump-start stage.  
We conduct two sets of experiments with $M = 2$, where one auxiliary dataset has a small $\varepsilon_m=2$ (low-quality data), and the other has a large $\varepsilon_m=8$ (high-quality data). The only difference between the two experiments lies in which of the two auxiliary datasets enters the jump-start stage first.
In Figure \ref{fig:order}, a significant performance gap on the target data is observed between starting with high-quality auxiliary data versus starting with low-quality data. 
We believe this gap arises due to arms that were mistakenly removed by low-quality auxiliary data. 
In particular, the algorithm can sometimes be overly aggressive in eliminating arms during the jump-start stage, which may incorrectly remove the optimal arm, leading to persistent regret in that area for the target data. 
These results suggest that starting with high-quality auxiliary data is recommended for achieving better overall performance.

\subsection{Real Data Experiments}\label{sec:realexperiments}
In this section, we further examine the performance of the proposed algorithms on three widely used classification datasets, whose summary statistics are given in Table \ref{tab:realdatainfo}. The detailed information for each dataset, including covariates, responses, pre-processing and selection of target and auxiliary data, are collected in Section \ref{sec:datasetdetails} of the supplement.

\begin{table}[]
\centering
\caption{Summary of real datasets.}
\resizebox{0.66\linewidth}{!}{
\renewcommand{\arraystretch}{0.9}
\setlength{\tabcolsep}{5pt}
\begin{tabular}{r|r|r|r|r|r|r}
\toprule
\multicolumn{1}{c|}{} & \multicolumn{1}{c|}{$n_{\mathrm{P}}$} & \multicolumn{1}{c|}{$M$} & \multicolumn{1}{c|}{$\max_m n_{\mathrm{Q}_m}$} & \multicolumn{1}{c|}{$K$} & \multicolumn{1}{c|}{\begin{tabular}[c]{@{}c@{}}original\\ dimension\end{tabular}} & \multicolumn{1}{c}{\begin{tabular}[c]{@{}c@{}}$d$ after\\ preprocessing\end{tabular}}  \\ \midrule
\textsc{Adult}         & 41292                                 & 7                        & 3930                                           & 2                        & 46                                                                                & 3                                                                                 \\ 
\textsc{Jobs}          & 57773                                 & 1                        & 14318                                          & 2                        & 11                                                                                & 3                                                                          \\ 
\textsc{Taxi}          & 621957                                & 1                        & 18945                                          & 2                        & 93                                                                                & 3                                                                                 \\ \bottomrule
\end{tabular}
}
\label{tab:realdatainfo}
\end{table}


In particular, we adopt the framework of creating bandit instances from (offline) classification datasets following \cite{riquelme2018deep} and \cite{dimakopoulou2019balanced}. Suppose we have a classification dataset $\{X_i, \dot{Y}_i\}_{i=1}^{n_{\mathrm{P}}}$, where the class labels $\dot{Y}_i \in [K]$. We regard the $K$ classes as the bandit arms and define the reward of the $k$-th arm as $Y_i^{(k)}=\mathbf{1}(\dot{Y}_i=k)$.
Let the underlying true relationship between $\dot{Y}$ and $X$ be $\dot{f}_k(X) :=  \mathbb{P}\big[\dot{Y} = k | X\big]$ for $k\in[K]$. This implies that the expected reward function of the $k$-th arm can be computed as 
\begin{align*}
f_{k}(x) :=\mathbb{E}\left[Y_i^{(k)}| X_i=x\right] =  \dot{f}_k(x), \text{ for all } k\in [K].
\end{align*}
Thus, if the class probability functions are smooth, the reward function $f_{k}$ is also smooth.

The evaluation metric is defined as the cumulative reward $\sum_{i = 1}^{n_{\mathrm{P}}}{Y}_i^{\pi_i(X_i)}$, with an expectation 
\begin{align*}
\mathbb{E}_{X, \dot{Y} } \left[\sum_{i=1}^{n_{\mathrm{P}}}{Y}_i^{\pi_i(X_i)}\right] = \sum_{i=1}^{n_{\mathrm{P}}}
\mathbb{E}_{X}\left[\sum_{k=1}^K \dot{f}_k(X_i) \mathbf{1}(\pi_i(X_i)=k)\right] 
= \sum_{i=1}^{n_{\mathrm{P}}} \mathbb{E}_X\left[ f_{\pi_i(X_i)}(X_i)\right],
\end{align*}
which is compatible with the regret defined in \eqref{equ:def-regret}. Note that since the true class probability functions $\{\dot f_k(\cdot)\}$ are \textit{unknown}, we cannot directly compute the reward as $\sum_{i = 1}^{n_{\mathrm{P}}}f_{\pi_i(X_i)}(X_i)$.


We consider three competing methods and a benchmark method:
\begin{itemize}
\item \texttt{LDPMAB}: our proposed method for LDP contextual nonparametric multi-armed bandits. We implement \texttt{LDPMAB} with and without (marked as w and wo, respectively) auxiliary data. 
\vskip -1mm

\item \texttt{Linear}: the method proposed in \cite{han2021generalized} for LDP contextual generalized linear bandits (see Algorithm 2 therein), which does not consider transfer learning. We set the parametric model for the expected reward of each arm as a logistic function. We also test the method with auxiliary data, where we include auxiliary data in the stochastic gradient descent of the parameter estimation with the required privacy level. 
\vskip -1mm

\item \texttt{NN}: we generalize \texttt{Linear} by replacing the expected reward model for each arm with a single-layer neural network, with the other steps staying unchanged.
\vskip -1mm

\item \texttt{ABSE}: the method proposed in \cite{perchet2013multi} for non-private contextual nonparametric multi-armed bandits, which does not consider transfer learning.
\end{itemize}
The implementation details of all methods can be found in \Cref{sec:implementationdetail} of the supplement and we present the experiment result based on 100 repetitions. To proceed, we first explain how the experiment is implemented for each repetition (for simplicity of presentation, we assume $M=1$). In particular, given the original target data $\{X_i^{\mathrm{P}}, \dot{Y}_i^{\mathrm{P}}\}_{i=1}^{n_{\mathrm{P}}}$ and auxiliary data $\{X_i^{\mathrm{Q}}, \dot{Y}_i^{\mathrm{Q}}\}_{i=1}^{n_{\mathrm{Q}}}$ from the (offline) classification dataset, the following steps are executed sequentially:
\begin{itemize}
    \item We first conduct a random permutation of the index $\{1,2,\cdots,n_{\mathrm{P}}\}$ and $\{1,2,\cdots,n_{\mathrm{Q}}\}$. With an abuse of notation, we denote the permuted data via $\{X_i^{\mathrm{P}}, \dot{Y}_i^{\mathrm{P}}\}_{i=1}^{n_{\mathrm{P}}}$ and $\{X_i, \dot{Y}_i^{\mathrm{Q}}\}_{i=1}^{n_{\mathrm{Q}}}$ as well.
    \item We now generate the bandit auxiliary data. For each $i\in [n_{\mathrm{Q}}]$, given $X_i^{\mathrm{Q}}$, we implement the behavior policy $\pi^{\mathrm{Q}}$, pull arm $\pi^{\mathrm{Q}}(X_i^{\mathrm{Q}})$ and observe the reward $Y_i^{\mathrm{Q}, (\pi^{\mathrm{Q}}(X_i^{\mathrm{Q}}))}:=\mathbf{1}(\dot Y_{i}^{\mathrm{Q}}=\pi^{\mathrm{Q}}(X_i^{\mathrm{Q}}))$. We thus attain the bandit auxiliary data $\mathcal{D}^{\mathrm{Q}}=\{Z_i^{\mathrm{Q}}\}_{i=1}^{n_{\mathrm{Q}}}$ where $Z_i^{\mathrm{Q}}=(X_i^{\mathrm{Q}}, \pi^{\mathrm{Q}}(X_i^{\mathrm{Q}}), Y_i^{\mathrm{Q}, (\pi^{\mathrm{Q}}(X_i^{\mathrm{Q}}))}).$
    \item For each of the four methods (i.e.\ \texttt{LDPMAB}, \texttt{Linear}, \texttt{NN}, \texttt{ABSE}), we now start the learning process on the target data for $i\in[n_{\mathrm{P}}]$, where note that given the pulled arm $\pi_i(X_i^{\mathrm{P}})$, the reward is generated via $Y_i^{\mathrm{P}, (\pi_i(X_i^{\mathrm{P}}))}:=\mathbf{1}(\dot Y_{i}^{\mathrm{P}}=\pi_i(X_i^{\mathrm{P}}))$. The cumulative reward is therefore $\sum_{i = 1}^{n_{\mathrm{P}}}{Y}_i^{\pi_i(X_i)}$.
\end{itemize}
Note that all three steps above involves randomness, stemming from permutation, realization of behavior policy, the privacy mechanism (i.e.\ Laplacian random noises), and realization of target policy.


The experiment results for each method~(\texttt{LDPMAB}, \texttt{Linear}, \texttt{NN}) on the three datasets are summarized in Table \ref{tab:realdataresult} under various combinations of privacy budgets $(\varepsilon,\varepsilon_m)$. Note that to standardize the scale across datasets, we report the ratio of the mean reward of each method relative to that of \texttt{ABSE}, which, as discussed above, is implemented on the target data non-privately without transfer learning. Thus, a reported value larger than $1$ means that the method is better than \texttt{ABSE} and {vice versa}. 

Several observations are in order. 
First, \texttt{LDPMAB} with auxiliary data outperforms its competitors in terms of both best performance (number of significantly better rewards) and average performance (rank-sum). 
This shows that our proposed methods can effectively utilize auxiliary data and thus achieves knowledge transfer with the designed jump-start scheme. 
In contrast, \texttt{Linear} and \texttt{NN} occasionally have negative transfer, where auxiliary data worsens the performance. In addition, without auxiliary data, \texttt{LDPMAB} still outperforms \texttt{Linear}, suggesting the advantage of the nonparametric nature of \texttt{LDPMAB}. Compared to \texttt{ABSE}, the competing methods without auxiliary data are usually worse (i.e.\ with ratio less than 1) since LDP is required, indicating the cost of privacy.



\begin{table}[!t]
\caption{The best performer among 6 methods (i.e.\ \texttt{LDPMAB} w/wo, \texttt{Linear} w/wo, \texttt{NN} w/wo) are marked in \textbf{bold} for each dataset under different combinations of $(\varepsilon, \varepsilon_m)$. Note that for each dataset, we report the performance at both $t = n_{\mathrm{P}} / 4$ and $t = n_{\mathrm{P}} $ to highlight the effect of transfer learning. To ensure statistical significance, we adopt the Wilcoxon signed-rank test  \citep{wilcoxon1992individual} with a significance level of 0.05 to check if the result is significantly better.
The best results that hold significance towards the others are \colorbox{lightgray}{highlighted} in grey. } 
\label{tab:realdataresult}
\centering
\resizebox{1\linewidth}{!}{
\renewcommand{\arraystretch}{0.99}
\setlength{\tabcolsep}{4pt}
\begin{tabular}{|c|cccccc|cccccc|}
\toprule
\multirow{3}{*}{Dataset} & \multicolumn{6}{c|}{$t =  n_{\mathrm{P}} / 4$}                                    & \multicolumn{6}{c|}{$t = n_{\mathrm{P}}$}                                         \\ \cmidrule{2-13} 
& \multicolumn{2}{c}{\texttt{LDPMAB}} & \multicolumn{2}{c}{\texttt{Linear}} & \multicolumn{2}{c|}{\texttt{NN}} & \multicolumn{2}{c}{\texttt{LDPMAB}} & \multicolumn{2}{c}{\texttt{Linear}} & \multicolumn{2}{c|}{\texttt{NN}} \\
& w            & wo          & w            & wo          & w          & wo         & w            & wo          & w            & wo          & w          & wo         \\ 
\midrule
\multicolumn{13}{|c|}{$(\varepsilon, \varepsilon_m) = (1, 1)$}
\\
\midrule 
\textsc{Adult}           & \textbf{1.459} \cellcolor[rgb]{0.8,0.8,0.8}       & 0.987       & 0.954        & 0.950       & 0.906      & 0.935      & \textbf{1.101}  \cellcolor[rgb]{0.8,0.8,0.8}       & 0.795       & 0.724        & 0.715       & 0.750      & 0.784      \\
\textsc{Jobs}            & \textbf{0.801}        & 0.797       & 0.786        & 0.794       & 0.800      & 0.795      & 0.694        & 0.693       & 0.684        & 0.690       & \textbf{0.700}      & 0.695      \\
\textsc{Taxi}            & 0.989        & 0.976       & 0.987        & 0.984       & \textbf{0.998}      & 0.989      & 0.994        & 0.992       & 0.996        & 0.993       & \textbf{0.998}      & 0.995      \\ \midrule

\multicolumn{13}{|c|}{$(\varepsilon, \varepsilon_m) = (1, 4)$}
\\

\midrule
\textsc{Adult}           & \textbf{1.602}  \cellcolor[rgb]{0.8,0.8,0.8}       & 0.987       & 0.988        & 0.986       & 1.110      & 0.992      & \textbf{1.210}   \cellcolor[rgb]{0.8,0.8,0.8}      & 0.795       & 0.782        & 0.771       & 0.871      & 0.816      \\
\textsc{Jobs}            & \textbf{0.846} \cellcolor[rgb]{0.8,0.8,0.8}        & 0.797       & 0.795        & 0.804       & 0.800      & 0.798      & \textbf{0.742}  \cellcolor[rgb]{0.8,0.8,0.8}       & 0.693       & 0.688        & 0.704       & 0.698      & 0.695      \\
\textsc{Taxi}            & \textbf{0.997}   \cellcolor[rgb]{0.8,0.8,0.8}      & 0.985       & 0.976        & 0.969       & 0.990      & 0.989      & \textbf{0.996}        & 0.992       & 0.992        & 0.991       & \textbf{0.996}      & 0.995      \\ \midrule

\multicolumn{13}{|c|}{$(\varepsilon, \varepsilon_m) = (2, 1)$}
\\
\midrule
\textsc{Adult}           & \textbf{1.459}  \cellcolor[rgb]{0.8,0.8,0.8}       & 0.986       & 0.964        & 0.986       & 0.895      & 0.930      & \textbf{1.102}    \cellcolor[rgb]{0.8,0.8,0.8}     & 0.919       & 0.762        & 0.772       & 0.745      & 0.808      \\
\textsc{Jobs}            & \textbf{0.819}   \cellcolor[rgb]{0.8,0.8,0.8}      & 0.808       & 0.788        & 0.791       & 0.800      & 0.797      & 0.719        & \textbf{0.720}  \cellcolor[rgb]{0.8,0.8,0.8}      & 0.683        & 0.683       & 0.705      & 0.688      \\
\textsc{Taxi}            & 0.992        & 0.974       & 0.989        & 0.989       & \textbf{1.000}      & 0.996      & 0.996        & 0.992       & 0.997        & 0.997       & \textbf{1.000}      & 0.999      \\ \midrule
\multicolumn{13}{|c|}{$(\varepsilon, \varepsilon_m) = (2, 4)$}
\\
\midrule
\textsc{Adult}           & \textbf{1.602} \cellcolor[rgb]{0.8,0.8,0.8}        & 0.986       & 0.964        & 0.968       & 0.895      & 0.929      & \textbf{1.210}  \cellcolor[rgb]{0.8,0.8,0.8}       & 0.919       & 0.762        & 0.762       & 0.745      & 0.791      \\
\textsc{Jobs}            & \textbf{0.857}  \cellcolor[rgb]{0.8,0.8,0.8}       & 0.808       & 0.788        & 0.785       & 0.800      & 0.792      & \textbf{0.754} \cellcolor[rgb]{0.8,0.8,0.8}        & 0.720       & 0.683        & 0.674       & 0.705      & 0.696      \\
\textsc{Taxi}            & \textbf{1.001}        & 0.974       & 0.989        & 0.989       & 1.000      & 1.000      & \textbf{1.002}        & 0.992       & 0.997        & 0.997       & 1.000      & 1.000      \\ \midrule

Rank sum  & \textbf{15} & 45 & 55 &  52 & 
37 & 43 & \textbf{22} & 44 & 54 & 56  & 33 & 36
\\ \bottomrule
\end{tabular}
}
\end{table}

\section{Conclusions and Discussions}\label{sec:discussionconclusion}

In this work, we investigate the problem of nonparametric contextual multi-armed bandits under local differential privacy.
We propose a novel uniform-confidence-bound based algorithm, which achieves near-optimal performance supported by a newly derived minimax lower bound. 
To further improve the performance limit of LDP contextual MAB, we consider transfer learning, which incorporate side information from auxiliary datasets that are also subject to LDP constraints. 
Assuming covariate shift, we introduce a jump-start scheme to leverage the auxiliary data, attaining the established minimax lower bound, up to logarithmic factors in interesting regimes. 
Extensive experiments on synthetic and real datasets validate our theoretical findings and demonstrate the superiority of our methodology. 

We remark on the implications of our method in the context of multi-task learning.
Consider a scenario where a set of $M$ players are deployed to engage in a bandit game, with the overall objective being to minimize the average regret across all players \citep{deshmukh2017multi, wang2021multitask}. These players simultaneously interact with a shared set of arms. At each round, each player selects an arm and receives feedback. 
The conditional distribution of each arm's reward is identical across all players.
Under this setting, the estimator in \eqref{equ:privateestimatormulti} is permutation invariant with respect to the datasets. This means that treating any dataset as the target dataset does not affect the estimator's effectiveness or the subsequent confidence bound \eqref{equ:rkj}.
This observation suggests that the proposed methodology can be extended to multi-task learning, provided Algorithm \ref{alg:ldpmab} is adapted to accommodate parallel interactions. We leave a thorough investigation for future research. 

\vspace{2mm}
\setlength{\bibsep}{1pt plus 1ex}
\begin{spacing}{1.5}
	\bibliographystyle{apalike}
	\bibliography{mul_stat}
\end{spacing}

\end{document}